\documentclass[sigconf]{acmart}

\copyrightyear{2022}
\acmYear{2022}
\setcopyright{acmcopyright}\acmConference[MM '22]{Proceedings of the 30th ACM
International Conference on Multimedia}{October 10--14, 2022}{Lisboa, Portugal}
\acmBooktitle{Proceedings of the 30th ACM International Conference on Multimedia
(MM '22), October 10--14, 2022, Lisboa, Portugal}
\acmPrice{15.00}
\acmDOI{10.1145/3503161.3547900}
\acmISBN{978-1-4503-9203-7/22/10}
\usepackage{amsmath,amsfonts,bm}

\def\eqref#1{equation~\ref{#1}}
\def\1{\bm{1}}

\def\vf{{\bm{f}}}

\def\vp{{\bm{p}}}
\def\vq{{\bm{q}}}

\def\vv{{\bm{v}}}
\def\vw{{\bm{w}}}

\def\vy{{\bm{y}}}

\DeclareMathAlphabet{\mathsfit}{\encodingdefault}{\sfdefault}{m}{sl}
\SetMathAlphabet{\mathsfit}{bold}{\encodingdefault}{\sfdefault}{bx}{n}

\def\sB{{\mathbb{B}}}

\def\sI{{\mathbb{I}}}

\def\sX{{\mathbb{X}}}

\usepackage{graphicx}
\usepackage{amsmath}
\usepackage{amssymb}
\usepackage{booktabs}
\usepackage{makecell}
\usepackage{multirow}
\usepackage{subcaption}
\usepackage{enumitem}

\usepackage{graphicx}
\newcommand{\tablestyle}[2]{\setlength{\tabcolsep}{#1}\renewcommand{\arraystretch}{#2}\centering\footnotesize}

\newcommand{\etal}{\textit{et al}.~}

\newcommand{\ieno}{\textit{i}.\textit{e}.}

\newcommand{\egno}{\textit{e}.\textit{g}.}

\newcommand{\etcno}{\textit{etc}}

\newcommand{\tcr}{\textcolor{red}}

\begin{document}

\title{Meta Clustering Learning for Large-scale Unsupervised Person Re-identification}

\author{Xin Jin}
\authornote{Corresponding Author.}
\email{jinxin@eias.ac.cn}
\affiliation{
    \institution{
    Eastern Institute for Advanced Study
    }
    \country{}
}

\author{Tianyu He}
\email{deeptimhe@gmail.com}
\affiliation{
    \institution{
    Alibaba Group
    }
    \country{}
}

\author{Xu Shen}
\email{shenxu.sx@alibaba-inc.com}
\affiliation{
    \institution{
    Alibaba Group
    }
    \country{}
}

\author{Tongliang Liu}
\email{tongliang.liu@sydney.edu.au}
\affiliation{
    \institution{
    The University of Sydney
    }
    \country{}
}

\author{Xinchao Wang}
\email{xinchao@nus.edu.sg}
\affiliation{
    \institution{
    National University of Singapore
    }
    \country{}
}

\author{Jianqiang Huang}
\email{jianqiang.jqh@gmail.com}
\affiliation{
    \institution{
    Alibaba Group
    }
    \country{}
}

\author{Zhibo Chen}
\email{chenzhibo@ustc.edu.cn}
\affiliation{
    \institution{
    University of Science and Technology of China
    }
    \country{}
}

\author{Xian-Sheng Hua}
\email{huaxiansheng@gmail.com}
\affiliation{
    \institution{
    Alibaba Group
    }
    \country{}
}

\begin{abstract}
  Unsupervised Person Re-identification (U-ReID) with pseudo labeling recently reaches a competitive performance compared to fully-supervised ReID methods based on modern clustering algorithms. However, such clustering-based scheme becomes computationally prohibitive for large-scale datasets, making it infeasible to be applied in real-world application. How to efficiently leverage endless unlabeled data with limited computing resources for better U-ReID is under-explored. {In this paper, we make the first attempt to the large-scale U-ReID and propose a ``small data for big task'' paradigm dubbed Meta Clustering Learning (MCL). MCL only pseudo-labels a subset of the entire unlabeled data via clustering to save computing for the first-phase training.} After that, the learned cluster centroids, termed as meta-prototypes in our MCL, are regarded as a proxy annotator to softly annotate the rest unlabeled data for further {polishing} the model. {To alleviate the potential noisy labeling issue in the polishment phase,} we enforce two well-designed loss constraints to promise intra-identity consistency and inter-identity strong correlation. For multiple widely-used U-ReID benchmarks, our method significantly saves computational cost while achieving a comparable or even better performance compared to prior works.
\end{abstract}


%   <concept>
%       <concept_id>10010147.10010178.10010224.10010245.10010252</concept_id>
%       <concept_desc>Computing methodologies~Object identification</concept_desc>
%       <concept_significance>500</concept_significance>
%       </concept>
%  </ccs2012>

% \ccsdesc[500]{Computing methodologies~Object identification}







\maketitle

\section{Introduction}

Ubiquitous cameras generate innumerable pedestrian data every day. Due to the growing demands on person re-identification (ReID) and its expensive labeling cost, unsupervised person ReID (U-ReID)~\cite{fan2018unsupervised,li2018unsupervised,tang2019unsupervised,qi2019novel,yu2019unsupervised,yang2019patch,ding2020adaptive,zhai2020ad,jin2020global,jin2020snr,ge2020self,dai2021cluster,zhuang2021joint} has attracted increasing attention recently.

% \begin{figure}
%   \centerline{\includegraphics[width=0.78\linewidth]{figures/motivation_v6.pdf}}
%   \vspace{-2mm}
%     \caption{Motivation illustration for clustering-based unsupervised ReID: as the size of unlabeled data increases, the clustering process in previous works will (a) cost an intolerable computational resources in terms of memory and time costs, and (b) be more prone to be affected by pseudo label noise. Our work introduces a new meta-clustering learning to achieve a satisfactory U-ReID performance while simultaneously tackling these two challenges.}
% \label{fig:motivation}
% \vspace{-6mm}
% \end{figure}

\begin{figure}
  \centering
  \includegraphics[width=1.0\linewidth]{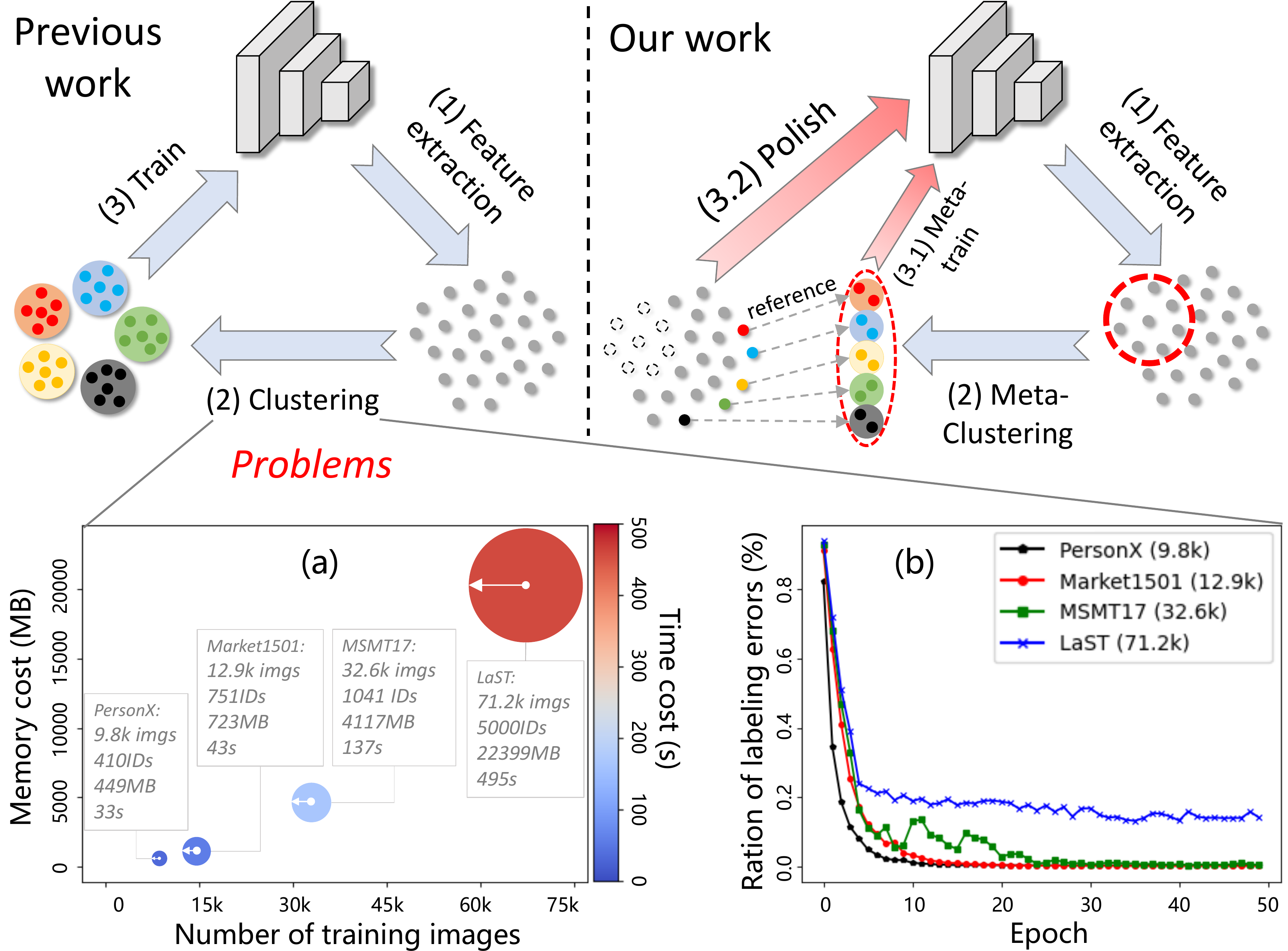}
  \caption{Motivation illustration for clustering-based unsupervised ReID: as the size of unlabeled data increases, the clustering process in previous works will (a) cost an intolerable computational resources in terms of memory and time costs, and (b) be more prone to be affected by pseudo label noise. Our work introduces a new meta-clustering learning to achieve a satisfactory U-ReID performance while simultaneously tackling these two challenges.}
  \label{fig:motivation}
\end{figure}

There are mainly two categories in U-ReID. One is unsupervised domain adaptive (UDA) person ReID, which first pre-trains a model on the labeled source dataset, and then fine-tunes the model on the unlabeled target dataset to reduce domain gap~\cite{deng2018image,lin2018multi,wang2018transferable,wei2018person,yu2019unsupervised,zhong2018generalizing,zhong2019invariance}. Albeit effective, UDA ReID branch typically suffers from a complex adaptation process, and its success also relies on an assumption that the discrepancy between source and target domain is not significant. This motivates the exploration on the other branch, the clustering-based unsupervised ReID~\cite{fan2018unsupervised,fu2019self,ge2020self,lin2019bottom,wang2020unsupervised,dai2021cluster}. As the ``Previous work'' shown in Figure~\ref{fig:motivation}, the works of this branch tend to perform an iterative optimization process of \emph{feature extraction--clustering--train}. In this way, all unlabeled data can be explicitly leveraged with the pseudo labels generated by clustering. The focus of the recent clustering-based methods lies in creating more reliable clusters and efficiently using them to learn discriminative representations, \egno, with the help of self-similarity grouping~\cite{fu2019self}, hybrid memory bank with contrastive loss~\cite{ge2020self} or cluster-level memory bank~\cite{dai2021cluster}, multi-label classification~\cite{wang2020unsupervised}, and online hierarchical cluster dynamics~\cite{zeng2020hierarchical,zheng2021online}.

% use a memory bank based architecture and contrastive loss

However, these methods all neglect an important fact in practice: the clustering process costs an intolerable computational resources due to its pair-wise similarity calculation and neighboring samples searching. Taking the most common clustering algorithm DBScan~\cite{ester1996density} in U-ReID as example, its worst time complexity and space complexity are both $O(n^2)$. When the size of unlabeled data is very large (as shown in Figure~\ref{fig:motivation}(a)), both of the memory and time cost of clustering will rapidly increase. {For example, performing clustering once on the LaST~\cite{shu2021large} (71.2k images) on the GPU following previous works~\cite{ge2020self,jin2020global,dai2021cluster} will take a memory usage up to 22GB , which can not run on a 16GB Tesla V100.} One may ask why not use the offline clustering (on CPU) or batch-wise local clustering (\egno, K-means) to avoid a large memory and time cost. This is due to the specificity of ReID: (1) the clustering-based ReID needs iteratively perform the feature extraction and clustering \textbf{in feature domain} (on GPU)~\cite{ge2020self,wang2020unsupervised,dai2021cluster}; (2) the batch-wise local clustering for ReID is sub-optimal, which hinders the exploration and utilization of global relationship among large-scale person data. 

% ``P$\times$K'' data sampling (L298) and loss construct.

% To {tackle} such limitation brought by clustering, there are two new directions are explored in recent works. One is unsupervised pre-training for ReID~\cite{fu2021unsupervised}, the other is semi-supervised ReID~\cite{he2021semi,ji2021meta}. However, the ReID pre-training aims to provide a better initialization weights for the downstream models. The semi-supervised ReID still relies on few labeled training samples for knowledge distillation. {These two branches} are not a purely unsupervised solution for large-scale ReID data.

In this paper, we attempt to achieve a large-scale unsupervised ReID framework while taking the computational cost into account, which is challenging but valuable and meaningful to bridge the gap between ReID algorithms and practical applications. To this end, we propose a ``small data for big task'' paradigm dubbed Meta Clustering Learning (MCL). Inspired by the other concept of Meta Learning~\cite{vilalta2002perspective,vanschoren2018meta,vanschoren2019meta} that are designed for `learning to learn' with the assistance of meta knowledge, our MCL first obtain the meta knowledge on a part of the unlabeled person data and then softly extend the knowledge to the rest unlabeled ones. Therefore, it naturally avoids clustering the full/whole dataset before each training epoch and thus reduces the computation overhead. In addition, during the knowledge extension process, MCL further leverages a clustering-free polishing step to enhance the discriminative representation learning while alleviating noisy label issue for ReID model.

% \tcr{Different from the popular concept of `meta learning' that focuses on "learning to learn"}, our MCL is an unified episodic training framework that clustering is only performed on \emph{part of the data} \tcr{to avoid clustering the full dataset at each epoch and thus reduce the computation overhead.} MCL also equips a clustering-free polishing step to enhance ReID model. MCL differs
% from existing U-ReID methods in exploiting unlabeled samples (illustrated in Figure~\ref{fig:motivation}) and exhibits advantages in computational cost saving and discriminative representation learning. This new task is challenging, but valuable and meaningful to bridge the gap between ReID algorithms and practical applications. 

% \tcr{One more point to be emphasized, here our `meta clustering learning (MCL)' is not equivalent to `meta-learning', where the former aims to use small data for big task while the latter focuses on "learning to learn".}

% ” paradigm where we first train ReID
% model with the computationally expensive clustering just
% on partial data so that saving resources, and then use the
% learned partial set as reference to annotate the rest unla-
% beled data for further polishing the mode}

{As illustrated in Figure~\ref{fig:motivation}, MCL consists of two phases of meta-prototype optimization and prototype-referenced polishment (see Sec.~\ref{sec:phase1},~\ref{sec:phase2} for details). In the first phase, the features of the \emph{partial} unlabeled images are extracted. This ratio can be flexibly determined according to the computing power of practical environment, as a by-product of MCL. Then, a clustering algorithm, like DBScan~\cite{ester1996density}, is used to cluster features and generate pseudo ID labels. Based on them, the ReID model is trained with a memory-based optimization strategy~\cite{ge2020self,wang2020unsupervised,dai2021cluster}. Meanwhile, the clustered centroids (termed as meta-prototypes) are stored in the memory and updated on the fly in a momentum manner~\cite{he2020momentum}.}

The second prototype-referenced {polishment} is based on the learned meta-prototypes in the previous phase, which are taken as a proxy annotator to mine the potential label information for \emph{the rest unlabeled data}. For each unused person image, we get a soft real-valued label likelihood vector by comparing it with meta-prototypes reference. Based on such clustering-free pseudo label, we further {polish} model by mining the relative comparative characteristic in person images. {The reason why we call this phase as ``polishment'' is because ``polish'' has the meaning of try to perfect one's skill, like here we promote the discriminative feature learning for ReID model with \emph{the rest unlabeled data}.}

Another point should be noticed is that, the pseudo labeling itself no matter of clustering-based or reference-based may generate wrong label predictions~\cite{wang2020unsupervised,jin2020global,wu2021mgh,ge2021cross}. As shown in Figure~\ref{fig:motivation}(b), the larger size of unlabeled dataset, the more possible of generating noisy labels. To alleviate it, we further leverage two loss constraints for label denoising in MCL. One loss enforces instance-level consistency to reduce intra-identity variance and the other constructs a soft-weighted triplet constraint to promise inter-identity correlation. In this way, MCL could better investigate the discriminative information of data even with noisy pseudo labels. {We
summarize our main contributions as follows:}

\begin{itemize}[leftmargin=*,noitemsep,nolistsep]

    \item To our best knowledge, this paper is the first to achieve the unsupervised large-scale ReID training while considering the computational cost savings. A ``small data for big task'' paradigm dubbed Meta Clustering Learning (MCL) is proposed. MCL performs clustering-based ReID training on \textbf{partial} unlabeled data, saving computing resources.
    
    \item To further leverage the rest unlabeled data, we take the learned prototypes from \textbf{partial} data as proxy annotator to pseudo-label them, and then polish model based on such pseudo labels with two well-designed losses (as a minor contribution) to promise intra-identity consistency and inter-identity strong correlation, which helps alleviate the noisy label issue.
    
    \item As the first attempt to handle the large-scale unsupervised ReID, extensive experiments on multiple benchmarks show that MCL could significantly save computational cost while achieving a state-of-the-art performance. In particular, MCL achieves ReID performance improvements of 4.8\%, 2.9\% in mAP on the large-scale MSMT17~\cite{wei2018person}, LaST~\cite{shu2021large}, but saves 71.8\%/87.9\% memory costs and 73.7\%/85.7\% time costs compared to the baselines.
    
\end{itemize}

\section{Related Work}

\subsection{Unsupervised Person Re-identification}

\noindent\textbf{Unsupervised Domain Adaptive (UDA) ReID.} This branch usually utilizes transfer learning, \egno, style translation~\cite{zhu2017unpaired}, to reduce domain gap between source and target ReID scenarios for adaptation~\cite{wei2018person,deng2018image,liu2019adaptive,dai2022bridging,zhang2022implicit}. Their performance is typically inferior to the clustering-based approaches, since there is still a gap between the style-translated images and the realistic person images~\cite{yang2019asymmetric,wang2020surpassing}.

\noindent\textbf{Clustering-based Unsupervised ReID.} This branch typically trains ReID model directly on unlabeled dataset with a clustering-based pseudo label estimation~\cite{song2020unsupervised,fan2018unsupervised,yang2019patch,zhang2019self,fu2019self,yang2019asymmetric,zheng2022clustering}. This clustering labeling and training process are usually alternatively performed until the model is stable. In particular, Lin~\etal~\cite{lin2019bottom} treat each individual sample as a cluster, and then gradually group similar samples into one cluster to generate pseudo labels. Jin~\etal~\cite{jin2020global} introduce a global distance-distribution separation constraint to handle the sample-wise noisy label. SPCL~\cite{ge2020self} proposes a self-paced contrastive learning framework to gradually create more reliable clusters for ReID training while updating the hybrid memory containing both source and target domain features. Similarly, ClusterContrast~\cite{dai2021cluster} further stores feature vectors inside a cluster-level memory to alleviate the inconsistent clustering issue. Recently, Isobe~\etal~\cite{isobe2021towards} introduce cluster-wise contrastive learning (CCL), progressive domain adaptation (PDA), Fourier augmentation (FA), and ICE~\cite{chen2021ice} introduces inter-instance contrastive encoding to boost the existing class-level contrastive ReID methods. However, all these methods focus on how to get more reliable pseudo labels or how to better leverage them for discriminative feature learning, an important point of computational cost is still under-explored. Besides, the noisy pseudo label issue in these methods has not yet been well addressed.

\begin{figure*}
  \centerline{\includegraphics[width=0.6\linewidth]{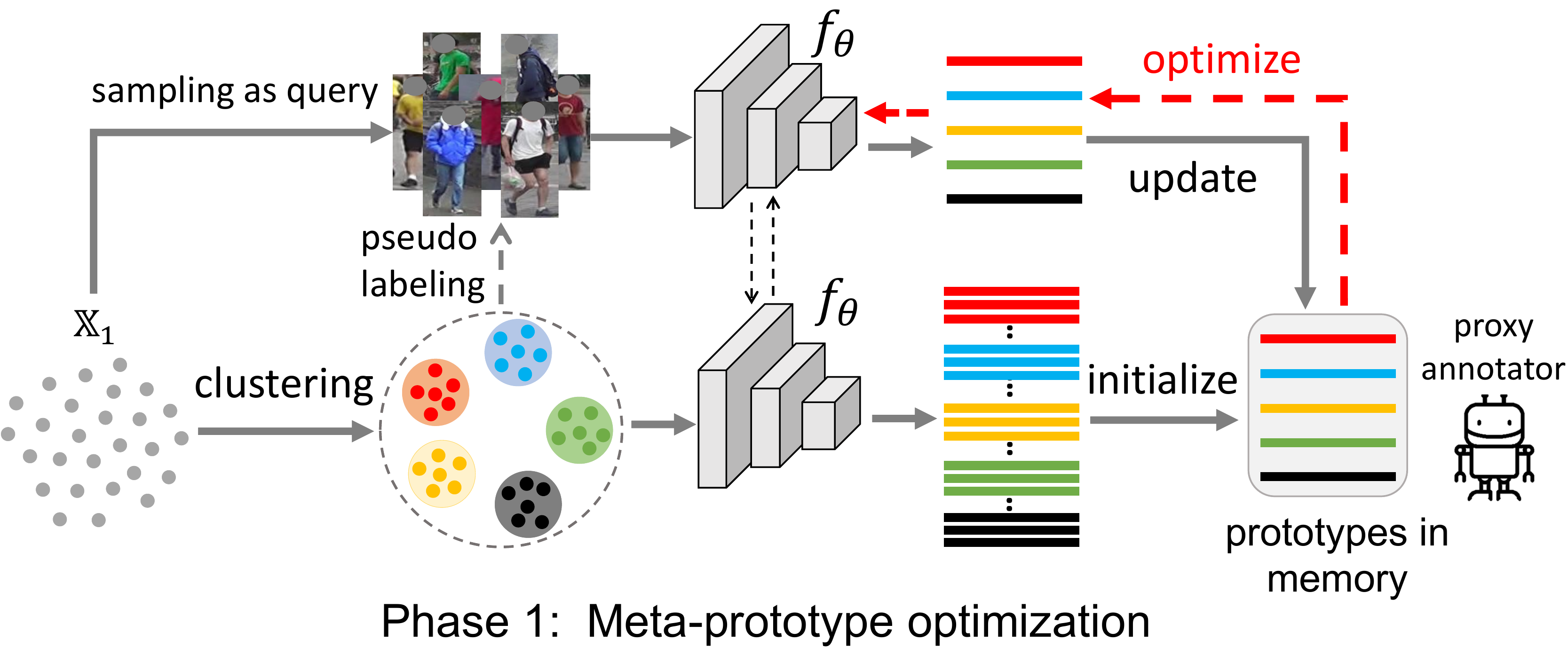}}
    \caption{The first training phase {in one training epoch} of MCL. Features of \emph{partial} data (meta-training subset $\sX_1$) are assigned pseudo labels by clustering, and the same color represents the same class. The lower part is the memory initialization while the upper part is the model training. A contrastive loss is calculated for optimization between query features and all prototypes in memory.}
\label{fig:phase1}
\vspace{-1mm}
\end{figure*}

\noindent\textbf{Reference-based Pseudo Labeling in ReID.} Existing representative works, like MAR~\cite{yu2019unsupervised}, MMCL~\cite{wang2020unsupervised}, MPRD~\cite{ji2021meta}, and SSL~\cite{lin2020unsupervised} either use labeled source data for pseudo labels generation or assign each unlabeled person image with a multi-class/softened label via pairwise similarity computation. Differently, our MCL creates clustering-free soft pseudo labels with the reference of online updated meta-prototypes that stored in the memory. Such design is more efficient because it does not need source labeled dataset as reference, and meta-prototypes (like a FC layer) could help directly infer out real-valued labels instead of repeat pairwise comparisons. Moreover, more reliable meta-prototypes encourage more accurate pseudo labeling (more effective unsupervised training), and vice versa. They promote each other, achieving a win-win effect. 

\subsection{Self-supervised Representation Learning}

% State-of-the-art methods on unsupervised visual representation learning [30,
% 45, 15, 41, 59, 13, 2] are based on the contrastive learning. Being cast as either the dictionary look-up
% task [45, 13] or the consistent learning task [41, 2], a contrastive loss was adopted to learn instance
% discriminative representations by treating each unlabeled sample as a distinct class.

% .

% Thus, SwAV~\cite{caron2020unsupervised} explores to take advantage of contrastive methods without requiring to compute pairwise comparisons. Simply put, it use a ``swapped'' prediction mechanism

MCL is also related to self-supervised representation learning (SSL). Based on contrastive learning framework, SSL has achieved a great success~\cite{ren2022shunted,feng2022image}, \egno, MoCo~\cite{he2020momentum}, MoCov2~\cite{chen2020improved}, SimCLR\cite{chen2020simple}, SimCLRv2~\cite{chen2020big}, BYOL~\cite{grill2020bootstrap}, and SimSiam~\cite{chen2021exploring}. Their main idea is to match a same instance in different augmented views, which typically relies on a large number of explicit pairwise feature comparisons and faces a computational challenge. Besides, these \textbf{instance-wise} SSL methods can not directly address the \textbf{fine-grained} unsupervised ReID problem (they can only be taken for pre-training/initialization~\cite{fu2021unsupervised,yang2021unleashing,fu2022large}), because ReID needs the cluster priors to mine fine discriminative clues.

\section{Meta Clustering Learning (MCL)}

\noindent\textbf{Overview.} To tackle the computing challenge in large-scale U-ReID, we propose a meta clustering learning (MCL), which is a unified episodic training framework, and comprises two phases of meta-prototype optimization (Figure~\ref{fig:phase1}) and prototype-referenced polishment (Figure~\ref{fig:phase2}). MCL alternates between these two phases: (1) group the \emph{partial} unlabeled data into clusters and store the learned meta-prototypes, while training model with cluster-level contrastive loss (Section~\ref{sec:phase1}); (2) use meta-prototypes as reference to annotate the \emph{rest} unlabeled samples for further fine-tune, and two loss constraints are enforced to promise intra-identity consistency and inter-identity correlation (Section~\ref{sec:phase2}). 

% , ~\ref{sec:noisy_label}

% We will introduce the details in the following sections.

% , two 
% square boxes in Fig.~\ref{fig:phase2}

% The main innovation of the proposed MCL framework lies in taking the limited computing resources to successfully leverage a large-scale unlabeled pedestrian data for ReID training. Its core idea is to first use a part of data for clustering-based 
% ReID training to get person identity prototypes, and then take these learned meta-prototypes as proxy annotator for pseudo labeling for fine-tuning with the rest unlabeled data. In order to avoid training error amplification caused by noisy labeling, the second fine-tune procedure additionally employs two novel loss constraints for optimization with considering intra-identity consistency and inter-identity \emph{relative} correlation (two 
% square boxes in Fig.~\ref{fig:phase2}).

% As mentioned above, our MCL scheme employs an episodic training strategy and alternates between two phases: (1) grouping the \emph{partial} unlabeled samples into clusters and storing the updated identity prototypes vectors in the memory dictionary, while training models with contrastive InfoNCE loss (Section~\ref{sec:phase1}), and (2) fine-tuning models by using meta-prototypes to annotating the \emph{rest} unlabeled samples for further training, in which two loss constraints that promises intra-identity consistency and inter-identity correlation are enforced. (Section~\ref{sec:phase2}). We will introduce the details sequentially in the following sections.

Given an unlabeled dataset $\sX$, MCL first splits $\sX$ into $N$ subsets uniformly, and then randomly selects one as \emph{meta-training subset} $\sX_{1}$ for meta-prototype optimization, \emph{the rest subsets} $\sX_{2},\sX_{3},...,\sX_{N}$ are taken for prototype-referenced polishment. This split is performed before each training epoch.

\subsection{Phase 1: Meta-prototype Optimization}\label{sec:phase1}

%  without any ground-truth label

%  based on clustering technique

% For example, the whole unlabeled training set can be divided into $N$ parts, including the meta training subset $\sX_{1}$ that taken for meta-prototype construction and optimization based on clustering technique, and the rest fine-tune subsets $\sX_{2},\sX_{3},...,\sX_{N}$ that taken for prototype-referenced fine-tune.

% , and then utilize the clustered pseudo labels for ReID training, which, in our opinion, is a sub-optimal solution because they all ignore the clustering operation may encounter computational bottleneck when meeting large-scale unlabeled dataset. 

% State-of-the-art U-ReID methods~\cite{ge2020self,dai2021cluster,isobe2021towards} rudely use all unlabeled data for clustering and training, which ignores the computational bottleneck issue of large-scale dataset. Instead, our 

MCL costs less resources in clustering by only using a meta-training subset $\sX_{1}$.

% in the first phase for training.

%  the SOTA U-ReID methods of

% to extract feature vectors, and then use a commonly-used clustering algorithm, such as DBScan~\cite{ester1996density} or K-means~\cite{macqueen1967some} to generate pseudo labels. Formally, 

% , we first apply the pre-trained weights on ImageNet into the neural network for initialization. 

\noindent\textbf{Feature Extraction and Clustering.} As shown in Figure~\ref{fig:phase1}, a network $f_{\theta}$ (\egno, ResNet-50~\cite{he2016deep}, initialized with pre-trained weights on ImageNet~\cite{ge2020mutual,ge2020self,dai2021cluster,isobe2021towards}) is taken as backbone to extract features from $\sX_{1}$. Then, DBScan~\cite{ester1996density} is used to cluster these features (unclustered outliers are discarded~\cite{chen2021ice,dai2021cluster}). The \emph{IDs} of cluster results are assigned to unlabeled samples as the pseudo labels for training.

\noindent\textbf{Query Setup and Meta-prototype Initialization.} After obtaining clustered pseudo labels for $\sX_{1}$, we sample $P$ person identities and $I$ instances for each identity, to set up a mini batch with the size of $P \times I$. Different from works~\cite{song2020unsupervised,jin2020global} that directly use the instance-wise loss contraints (\egno, triplet loss~\cite{hermans2017defense}) for training, we take each batch as a query set and employ a memory dictionary based contrastive learning~\cite{ge2020self,dai2021cluster,isobe2021towards} for optimization.

% \noindent\textbf{Memory Initialization.}    performing forward computation of

% store all cluster centroids, termed as

% for the different training epochs

% The memory dictionary is initialized with the features extracted by $f_\theta$. 

% which is changing during the training as the clustering process will be performed in every epoch. The initial meta-prototypes are initialized with 

\begin{figure*}
  \centerline{\includegraphics[width=0.8\linewidth]{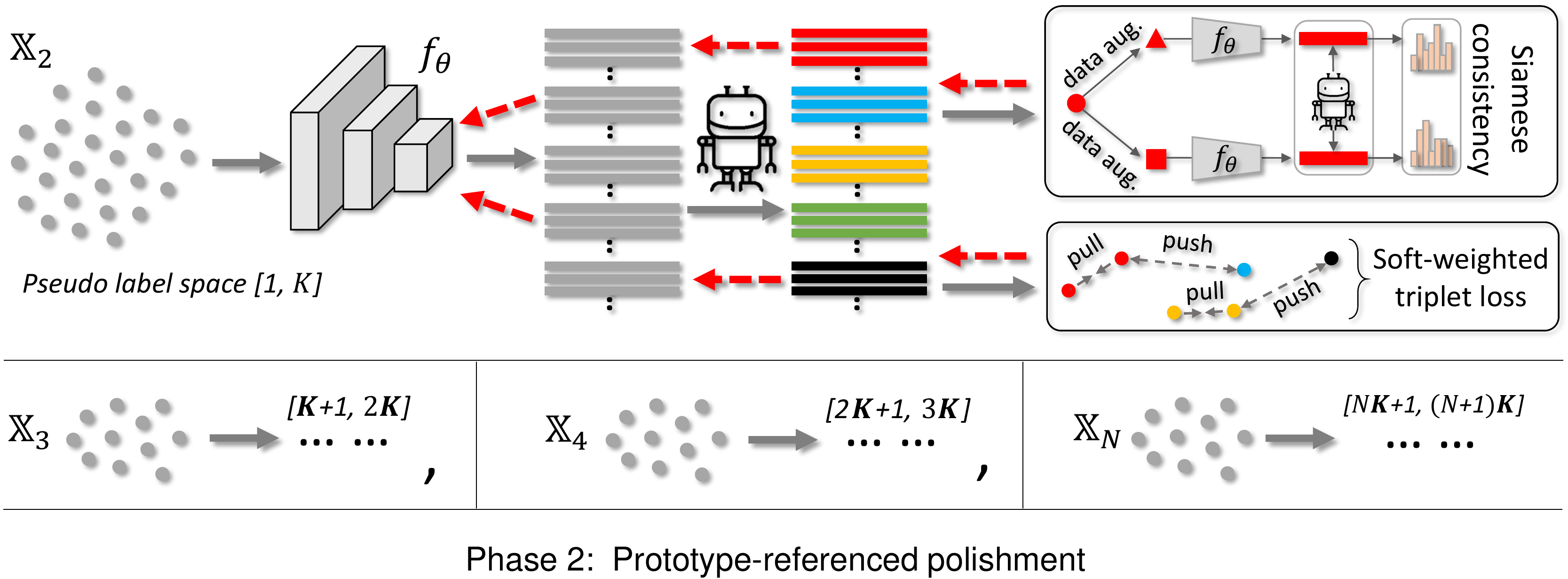}}
  \vspace{-2mm}
    \caption{The second phase in one training epoch of MCL. The rest unlabeled subsets $\sX_2,\sX_3,\cdots,\sX_N$ are assigned soft pseudo labels by comparing with the learned meta-prototypes (\ieno, the proxy annotator of `robot') for ReID model polishment. Siamese consistency loss $\mathcal{L}_{sc}$ and soft-weighted triplet loss $\mathcal{L}_{tri}^{sw}$ are enforced to alleviate the noisy pseudo label issue. Note that, this clustering-free polishment is performed on all subsets $\sX_2,\sX_3,\cdots,\sX_N$ together by cross-set sampling, with the non-overlapping label space.}
\label{fig:phase2}
\vspace{-1mm}
\end{figure*}

We maintain a group of learnable meta-prototypes $\{\vw_1, \cdots, \vw_K\}$ stored in the memory dictionary. Here, $K$ is same with the number of clustered clusters, {which is always changing during the training. Particularly, the clustering algorithm (\egno, DBScan) is performed before each training epoch, and then the \textbf{epoch-wise} meta-prototypes are initialized with} the mean feature vectors of each cluster, \ieno, $\vw_{k} = \frac{1}{|{\sI}_{k}|}\sum\vv_i$,
where $\vv_i$ means $i$-th feature vector of $k$-th cluster, ${\sI}_{k}$ denotes the $k$-th set that contains all the feature vectors within cluster $k$, and $|\cdot|$ denotes the number of features in the set.

% \limits_{\vv_i \in {\sI}_{k}} 

\noindent\textbf{Meta-prototypes Update and Model Optimization.} At each iteration $t$ of epoch, the encoded feature vectors $\{\vq\}$ of $P \times I$ query images in each mini-batch would be involved in meta-prototypes update. With the momentum updating~\cite{he2020momentum}, the $k$-th cluster prototype $\vw_k$ is updated by the mean of encoded query features belonging to class $k$,
\begin{align}
\label{eq:m_cen}
	\vw_k^t \leftarrow m \cdot \vw_k^{t-1} + (1-m)\cdot
	\frac{1}{|{\sB}_{k}^t|}\sum_{\vq^t_i \in {\sB}_{k}^t} \vq^t_i,
\end{align}
where $\sB_k^t$ denotes the feature vector set belonging to class $k$ in the mini-batch at the $t$-th iteration, and $m \in [0,1]$ is a momentum coefficient, which is empirically set as $0.2$ following~\cite{ge2020self,dai2021cluster}. The learned meta-prototypes are taken for model optimization together with query samples in this phase, and also play a role of proxy annotator (see the `robot' in Figure~\ref{fig:phase1},\ref{fig:phase2}) for the rest unlabeled subsets $\sX_{2},\sX_{3},...,\sX_{N}$ in the next phase.

% We take these stored and online updated cluster centroids in the memory dictionary as \emph{Meta-prototypes}, which are taken for model optimization in the first phase and are also deemed as proxy annotator (see the `robot' in Fig.~\ref{fig:phase1} and Fig.~\ref{fig:phase2}) for the rest unlabeled fine-tune subsets $\sX_{2},\sX_{3},...,\sX_{N}$.

With respect to the loss function in the first phase, we use a general contrastive loss~\cite{ge2020self,dai2021cluster} for model optimization. Basically, given a query instance $\vq$, we compare it to all meta-prototypes $\{\vw_1, \cdots, \vw_K\}$ using InfoNCE loss~\cite{oord2018representation}:
\begin{equation}\label{eq:phase1_loss}
    \mathcal{L}_{phase1} = -\log\frac{\exp(\vq \cdot \vw^+/\tau)}{\sum_{i=0}^K{\exp(\vq \cdot \vw_i/\tau)}}
\end{equation}
where $\vw^+$ is the positive cluster prototype vector to query instance $\vq$ and $\tau$ is a temperature hyper-parameter per~\cite{wu2018unsupervised}.

% The loss value is low when $q$ is similar to its positive cluster feature $c^+$ and dissimilar to all other cluster features.
% It is a log loss of K-way softmax-based classifier that tries to classify $q$ as $c^+$

% design
% a novel contrastive loss to fully exploit available data by treating all the source-domain classes,
% target-domain clusters and target-domain un-clustered instances as independent classes.

\subsection{Phase 2: Prototype-referenced Polishment}\label{sec:phase2}

% , we further adopt the clustering-free prototype-referenced fine-tune to leverage the rest unlabeled subsets $\sX_{2},\sX_{3},...,\sX_{N}$. 

To achieve the savings of clustering cost, the first meta-prototype optimization phase only uses a part of unlabeled data, \ieno, the meta-training subset $\sX_1$. The rest unlabeled subsets $\sX_{2},\sX_{3},...,\sX_{N}$ are leveraged in a clustering-free manner in this second phase. Basically, we take the learned meta-prototypes $\{\vw_1, \cdots, \vw_K\}$ from the first phase as proxy annotator to softly mine discriminative information for those rest unlabeled data for model polishment. This two-phase training is equivalent to traverse the entire dataset once, \ieno, one epoch.

%as prior works did.

% , no any increasing in computational cost

% Based on this clustering-free soft pseudo labels, we further fine-tune the ReID model by mining the relative comparative characteristic of different person images. 

\noindent\textbf{Prototype-referenced Labeling.} For clarity, we take an unused and unlabeled subset $\sX_2$ as example for illustration. Given $\sX_2 = \{x_i\}_{i=0}^{N_u}$ where each $x_i$ is a collected unlabeled person image, the learned meta-prototypes $\{\vw_1, \cdots, \vw_K\}$ defines the pseudo label space for $\sX_2$ is $[1, K]$. As shown in the `robot' in Figure~\ref{fig:phase2}, the meta-prototypes set $\{\vw_k\}_{k=1}^{K}$ is taken as proxy annotator, a soft real-valued pseudo label $\vy_i$ can be assigned for $x_i$ by comparing $f(x_i)$ with the reference agents $\{\vw_k\}_{k=1}^{K}$. This soft prototype-referenced pseudo labeling process is,
\begin{equation}\label{eq:labeling}
    y_i^j = L(f(x_i), \{\vw_k\}_{k=1}^{K})^j = \frac{\exp(\vw_j^{\rm T}f(x_i))}{\sum_{k}\exp(\vw_k^{\rm T}f(x_i))},
\end{equation}
where $L(\cdot)$ means a soft pseudo labeling function. {This function is epoch-wise and acts like a {dimension-variable} FC layer (\ieno, dot-product).} $y_i^j\in(0,1)$ is the $j$-th entry of $\vy_i$. All dimensions of $\vy_i$ add up to 1 and each dimension represents the label likelihood w.r.t. a reference prototype person ID. Different from the vanilla reference-based pseudo labeling~\cite{yu2019unsupervised} or {using a global classifier for labeling, our prototype-referenced labeling allows the \textbf{epoch-wise} and \textbf{dimension-variable} proxy annotator $\{\vw_k\}_{k=1}^{K}$ to be updated on the fly. More reliable meta-prototypes encourage more accurate labeling and thus more effective optimization, and vice versa. Besides, the clustering results are different ($K$ is changing) for each epoch in ReID, an immutable global classifier is infeasible.} 

% They promote and complement each other, achieving a win-win effect. 

% number of classi-
% fier parameters needs to be constantly changing (see L223).
% Q5: Comparison of the total training time

% In addition, following~\cite{wang2017normface,liu2017sphereface,yu2019unsupervised}, an unit norm constraint $||f(\cdot)||_2=1, ||\vw_k||_2=1, \forall i$ is enforced for a compact feature embedding. 

Last but not least, as shown in Figure~\ref{fig:phase2}, the label spaces for different subsets $\sX_{2},\sX_{3},...,\sX_{N}$ are manually designed as non-overlapping by considering two aspects: 1). the learned meta-prototypes $\{\vw_k\}_{k=1}^{K}$ can only cover a limited number of person identities ($K$ IDs) of the entire dataset, especially when the meta-training subset $\sX_{1}$ is very small. 2) assigning different subsets with different label spaces could increase the identity diversity, which simulates the real scenario where each person has multiple images from different views (see Sec.~\ref{sec:ablation} in experiment). 

% We have further discussed this in the experiment.

% a large-scale pedestrian dataset is typically collected from the different surveillance scenarios with numerous persons

% , and MCL taking the learned meta-prototypes as reference for labeling may inadvertently amplify such negative effect

% \begin{equation}\label{eq:harden_labeling}
%     J = \mathop{\arg\max} \ \ \{y^j\}_{j=1}^{K}.
% \end{equation}

\noindent\textbf{Polish the Model with Soft Pseudo Labels.} An intuitive solution is to `harden' the obtained soft pseudo labels, \ieno, regard the dimension with the largest value as person ID, \ieno, $J = \mathop{\arg\max} \ \ \{y^j\}_{j=1}^{K}$. Based on these identity labels $J$s, we can construct ReID constraints for training, such as cross-entropy loss~\cite{sun2018beyond,fu2019horizontal} and triplet loss~\cite{hermans2017defense}. In fact, our early attempts along this line have failed to deliver very good results. We analyze that is because the reference-based pseudo labeling itself will inevitably introduce some noisy labels. To exploit the merits of the prototype-referenced labeling while 
alleviating the noisy label effect, we additionally introduce two well-designed loss constraints to better leverage these \emph{unsatisfactory} soft pseudo labels for optimization, by considering the intra-identity consistency between different views within each person identity and the inter-identity correlations among different persons. They are shown in the two black boxes in Figure~\ref{fig:phase2} and elaborated below:

\noindent\textbf{Siamese consistency Loss $\mathcal{L}_{sc}$} is the first constraint to promise consistency within the same identity. As shown in the upper black box in Figure~\ref{fig:phase2}, $\mathcal{L}_{sc}$ is built on the ``swapped'' prediction idea of SwAV~\cite{caron2020unsupervised} to predict the label of a view
from the representation of another view. Given two features $\vf_s$ and $\vf_t$ extracted from two different augmentations of \emph{the same image}, we compute their referenced pseudo labels $\vy_s$ and $\vy_t$ following Eq.(\ref{eq:labeling}) by matching these features to the learned meta-prototypes $\{\vw_k\}_{k=1}^K$:
\begin{equation}
\begin{aligned}
  \mathcal{L}_{sc}(\vf_s, \vf_t) &= \mathcal{L}_{ce}(\vf_s, \vy_t) + \mathcal{L}_{ce}(\vf_t, \vy_s) \\
  \mathcal{L}_{ce}(\vf_s, \vy_t) &= - \sum_{k} \vy_t^{(j)}  \log \vp_s^{(j)}, 
  \hspace{3mm} \text{where} \hspace{2mm} \vp_s^{(j)} = \frac{\exp(\vw^\top_j \vf_s) }{\sum_{k} \exp(\vw_{k}^\top \vf_s) },
  \label{eq:loss}
\end{aligned}
\end{equation}
where $\mathcal{L}_{ce}(\vf_s, \vy_t)$ measures the fit between features $\vf_s$ and soft pseudo label $\vy_t$. Intuitively, if these two features capture the same person information, they should be able to predict from each other. $\mathcal{L}_{ce}$ is the cross entropy loss between the label and the probability obtained by taking a softmax of the dot products of $\vf$ and all prototypes in $\{\vw_k\}_{k=1}^K$.

% Formally, each term represents , where $\tau$ is a temperature parameter~\cite{wu2018unsupervised}.

% the same-identity samples and different-identity samples

% for the similarity measure of

\noindent\textbf{Soft-weighted Triplet Loss $\mathcal{L}_{tri}^{sw}$} is the second constraint which softly leverages the \emph{relative} correlations between identities to construct weighted-triplets for optimization. Considering the soft pseudo labels are continuous real-valued, a soft-weighted triplet loss $\mathcal{L}_{tri}^{sw}$ is enforced to promise the correct \emph{relative} correlation among person identities. Let $\{x^a, x^p, x^n\}$ be an input triplet sample and the corresponding feature embeddings are $\{f(x^a), f(x^p), f(x^n)\}$, the soft-weighted triplet loss is given by,

{\footnotesize
\begin{equation}
\begin{aligned}\label{eq:loss_tri}
\mathcal{L}_{tri}^{sw} = \omega(a,p,n)[\|f(x^a)&-f(x^p))\|_2^2 - \|f(x^a)-f(x^n))\|_2^2 + m]_{+}, \\
\omega(a,p,n) &= \left\langle f(x^a),f(x^p) \right\rangle \left\langle f(x^a),f(x^n) \right\rangle
\end{aligned}
\end{equation}
}%
where $\omega(a, p, n)$ and $m$ are the loss weighting factor and margin factor (0.3 by fault), $\left\langle\cdot\right\rangle$ means the similarities between feature vectors, which adaptively alters the magnitude of the triplet loss in a soft manner. In general, when the anchor-positive
pair is similar (\ieno, $\left\langle f(x^a),f(x^p) \right\rangle$ is high), the sample is more confident and reliable. Likewise, when the anchor-negative pair is similar (\ieno, $\left\langle f(x^a),f(x^n) \right\rangle$ is high), it forms \emph{a hard negative example}~\cite{schroff2015facenet}. Hence, $\mathcal{L}_{tri}^{sw}$ can give a higher priority and more attention on these reliable and hard cases, so as to alleviate the noisy label issue.

%   between feature vectors of anchor-positive and anchor-negative pairs
% \noindent\textbf{Overall Episodic Training Pipeline.} As illustrated in Fig.~\ref{fig:phase2}, the prototype-referenced fine-tune on $\sX_2,\sX_3,\cdots,\sX_N$, are conducted in parallel, with non-overlapping label space. 

\begin{comment}
% we first need to find a reliable metric to decide the identities of two person images are same or different. Inspired by a fact revealed by~\cite{yu2019unsupervised} that if a similar pair has highly similar comparative characteristics, it is probably a positive pair (otherwise, it is probably a hard negative pair), we employ a soft agreement metric $A(\cdot, \cdot)$ to measure the comparative characteristics encoded in a pair of prototype-referenced soft pseudo labels $(\vy_a, \vy_b)$,

% {\footnotesize
% \begin{align}\label{eq:agreement}
% A(\vy_a, \vy_b) = \vy_a \wedge \vy_b = \Sigma_k \min(\vy_a^{k}, \vy_b^{k}) =  1-\frac{||\vy_a-\vy_b||_1}{2},
% \end{align}
% }%
% based on L1 distance, this soft agreement is an analog to the voting by the meta-prototypes, each prototype $\vw_k$ in $\{\vw_k\}_{k=1}^K$ gives its conservative agreement $\min(\vy_a^{k}, \vy_b^{k})$ on believing the pair to be positive. In experiments, we empirically set an agreement threshold $A$ as $0.8$ to decide the positive-negative relationship for person images. 

% After that, 
\end{comment}

\begin{table}[bp]         
    \centering
    \footnotesize
    \caption{Introduction and comparison of datasets we used.}
    \vspace{-2mm}
    \resizebox{1.0\linewidth}{!}{
    \tablestyle{2.pt}{1.6}
    \begin{tabular}{c|ccccccc}
        \hline
        Dataset & Style & Train IDs & Train images & Test IDs & Query images & Total images & Cameras \\
        \hline
        
        PersonX~\cite{sun2019dissecting} & Synthetic & 410 & 9,840 & 856 & 5,136 & 45,792 & 6 \\
        
        Market-1501~\cite{zheng2015scalable} & Real & 751 & 12,936 & 750 & 3,368 & 32,668 & 6 \\
        
        MSMT17~\cite{wei2018person} & Real & 1,041 & 32,621 & 3,060 & 11,659 & 126,441 & 15 \\
        
        LaST~\cite{shu2021large} & Real & 5,000 & 70,923 & 5,803 & 10,173 & 228,156 & * \\
        
        \hline
    \end{tabular}}
    \label{tab:dataset}
    \vspace{-0mm}
\end{table}

% Table generated by Excel2LaTeX from sheet 'Sheet3'
\begin{table*}[htbp]
  \centering
  \footnotesize
  \caption{Memory\&Time Cost \textbf{\emph{vs.}} Unsupervised ReID Performance (\%). In which, $M (MB)$, $T (s)$ denotes the \emph{memory cost}, \emph{time cost} of performing clustering \textbf{once} in training, where `s' means `second'. $T (h)$ denotes the total training time where `h' means `hour'. We compare several MCL variants to baseline (\emph{All}, \ieno, Full Clustering scheme) by using \emph{50\%}, \emph{33\%}, \emph{25\%}, and \emph{20\%} data randomly selected from the entire unlabeled dataset as meta-training subset $\sX_{1}$. For the smallest dataset PersonX~\cite{sun2019dissecting}, it is not necessary to do experiments with too harsh computational requirements (\egno, \emph{33\%}, \emph{25\%}, \emph{20\%}). We can see that the larger size of unlabeled dataset, the more superior of our method (red). Note that, the DukeMTMC-ReID dataset~\cite{ristani2016performance} has been taken down and thus not used in our experiment, we just use PersonX~\cite{sun2019dissecting}, Market1501~\cite{zheng2015scalable}, MSMT17~\cite{wei2018person}, and LaST~\cite{shu2021large} for experiments.}
  \renewcommand\arraystretch{1.5}
  \vspace{-2mm}
  \setlength{\tabcolsep}{1.3mm}{
    \begin{tabular}{cccccc|ccccc|ccccc|ccccc}
    \toprule
    \multirow{2}[2]{*}{Methods} & \multicolumn{5}{c|}{PersonX (9.8k imgs, 410 IDs)} & \multicolumn{5}{c|}{Market1501 (12.9k imgs, 751 IDs)} & \multicolumn{5}{c|}{MSMT17 (32.6k imgs, 1041 IDs)}   & \multicolumn{5}{c}{LaST (71.2k imgs, 5000 IDs)} \\
\cmidrule{2-21}          & mAP   & Rank1 & M (MB)    & T (s)  & T (h)     & mAP   & Rank1 & M (MB)    & T (s)    & T (h)  & mAP   & Rank1 & M (MB)    & T (s)  & T (h)    & mAP   & Rank1 & M (MB)    & T (s) & T (h)  \\
    \hline
    
    All   & 88.5  & 95.8  & 822.3 & 30.0 & 2.7 & 83.3  & 93.0  & 876.3 & 34.3 & 2.9 & 33.4  & 62.9  & 6251.5 & 118.3 & 9.3 & 19.8  & 74.0  & 22398.5 & 494.8 & 42.0 \\
    
    50\%  & \textcolor[rgb]{0,  0, 1}{79.0}  & \textcolor[rgb]{0,  0, 1}{93.5}  & \textcolor[rgb]{0,  0, 1}{412.6} & \textcolor[rgb]{0,  0, 1}{13.1} & \textcolor[rgb]{0,  0, 1}{2.2} & 82.9  & 92.7  & 348.6 & 10.8 & 2.4 & \textcolor[rgb]{ 1,  0,  0}{38.2} & \textcolor[rgb]{ 1,  0,  0}{66.5} & \textcolor[rgb]{ 1,  0,  0}{1761.3} & \textcolor[rgb]{ 1,  0,  0}{31.1} & \textcolor[rgb]{ 1,  0,  0}{4.6} & \textcolor[rgb]{ 1,  0,  0}{20.0} & \textcolor[rgb]{ 1,  0,  0}{74.9} & \textcolor[rgb]{ 1,  0,  0}{5779.6} & \textcolor[rgb]{ 1,  0,  0}{121.2} & \textcolor[rgb]{ 1,  0,  0}{20.0}\\
    
    33\%  & --      & --      & --      & -- & -- & 79.6  & 91.9  & 287.5 & 7.0  & 2.2  & 31.5  & 57.4  & 889.1 & 18.2 & 3.8 & \textcolor[rgb]{ 1,  0,  0}{22.7}  & \textcolor[rgb]{ 1,  0,  0}{75.0}  & \textcolor[rgb]{ 1,  0,  0}{2688.2} & \textcolor[rgb]{ 1,  0,  0}{70.8} & \textcolor[rgb]{ 1,  0,  0}{14.0}\\
    
    25\%  & --      & --      & --      & -- & -- & 75.4  & 89.3  & 235.1 & 5.4  & 2.0  & 25.9  & 53.4  & 556.9 & 13.4 & 3.0 & 17.2  & 69.0  & 1564.2 & 44.7 & 9.0 \\
    
    20\%  & --      & --      & --      & -- & -- &  41.3  & 61.3  & 141.4 & 4.6  & 1.9   & 20.1  & 47.4  & 394.6 & 10.3 & 2.6 & 15.8  & 56.0  & 1066.0 & 38.4 & 7.0\\
    
    \bottomrule
    \end{tabular}}%
    \vspace{-1mm}
  \label{tab:table1}%
\end{table*}%

\vspace{-1.5mm}
\section{Experiment}

\subsection{Datasets and Implementation}

\noindent\textbf{Datasets and Evaluation.} We evaluate the proposed MCL method on multiple ReID benchmarks (from small to large scale): PersonX (\emph{PX})~\cite{sun2019dissecting}, Market1501 (\emph{Ma})~\cite{zheng2015scalable}, MSMT17 (\emph{MT})~\cite{wei2018person}, and the largest public ReID dataset (so far) LaST (\emph{LS})~\cite{shu2021large}. To further show the superiority of MCL in the large-scale data setting, we also conduct experiments on the \emph{mixed} datasets, \egno, training on multiple datasets \emph{PX+Ma+MT+LS} while testing on unseen test set of \emph{MT}. The details about datasets are shown in Table~\ref{tab:dataset}.

% are in \textbf{Supplementary}.

%  while further saving computational cost

\noindent\textbf{Implementation Details.} The proposed MCL is generic and can be applied to different clustering-based U-ReID backbones. Here, we re-implement ClusterContrast~\cite{dai2021cluster} as baseline, since it has been dominating the leaderboard in multiple benchmarks w.r.t unsupervised ReID performance, and is considerably more efficient as a source-free purely unsupervised ReID pipeline compared to those competitive adaptive (source data needed) U-ReID algorithms, like MMT~\cite{ge2020mutual}, SpCL~\cite{ge2020self}, ~\cite{zheng2021online}~\etcno. ResNet-50~\cite{he2016deep} is adopted as the backbone of the feature extractor and initialize the model with the parameters pre-trained on ImageNet~\cite{deng2009imagenet}. 

% With respect to the architecture, we make some modifications to the raw ResNet-50 to let it more suitable for ReID. Specifically, after layer-4, we remove all sub-module layers and add global average pooling (GAP) followed by batch normalization layer~\cite{ioffe2015batch} and L2-normalization layer, which will produce a feature vector with 2048-dimensions. During testing, we take the features of the global average pooling layer to calculate the distance for inference. 

% \noindent\textbf{Training and Hyper-parameters Setup.} 

At the beginning of MCL training, we first train the ReID model only with the first phase of \emph{meta-prototype optimization} (skip the second phase of \emph{prototype-referenced polishment}), which aims to warm up the \emph{meta-prototype} learning, like the FC layer warm-up~\cite{he2016deep,ro2019backbone}, so as to have a reasonable pseudo labeling for the next model polishing. This process lasted for 5 epochs for PersonX, and 10epochs for Market1501, MSMT17, and LaST. For image size, the input is resized as 256$\times$128 (height$\times$width) for all person datasets.
For data augmentation, we perform random horizontal
flipping, padding with 10 pixels, random cropping, and
random erasing~\cite{zhong2020random}. For batch size, each mini-batch contains 256 images of 16 pseudo person identities (16 instances for each person). During the training, we adopt Adam optimizer to train the ReID model with weight decay 5e-4. The initial learning rate is set to 3.5e-4, and is decayed by a factor of 0.1 every 20 epoch in a total of 60 epoch. Following~\cite{ge2020self,dai2021cluster}, we use DBScan and Jaccard distance~\cite{zhong2017re} 
to cluster with $k$ nearest neighbors, where $k$=30. For DBScan, the maximum distance $d$ between two samples is experimentally set as 0.4 for market1501, 0.7 for other datasets, and the minimal number of neighbors in a core point is all set as 4.

\subsection{Effectiveness and Necessity of MCL} \label{sec:table1}

%%achieve a good trade-off between computational efficiency and performance.
\noindent\textbf{Memory\&Time Cost \textbf{\emph{v.s.}} U-ReID Performance.} Table~\ref{tab:table1} shows the U-ReID performance resulted by using subsets of size \emph{50\%, 33\%, 25\%, and 20\%} randomly selected from all unlabeled data as meta-training set $\sX_{1}$ \emph{vs.} directly conducting clustering over full data (\emph{All}). We observe that using \emph{partial} data for clustering with MCL effectively saves computational costs on both memory and time. For example, the \emph{MCL, 50\%} schemes nearly achieve the same ReID performance but save memory\&time cost by over 50\%, such savings are particularly obvious on the large datasets: 1761.3MB, 31.1s vs 6251.5MB, 118.3s on MSMT17 and 5779.6MB, 121.2s vs 22398.5MB, 494.8s on LaST. {However, we also observe that the ReID performance of mAP/Rank1 has a noticeable drop, especially on the small dataset PersonX (`\textcolor{blue}{blue}' in Table~\ref{tab:table1}). We analyze that's because the noisy label issue will be enlarged on small datasets.}

Interestingly, in contrast to the
trend in Memory\&Time saving \emph{vs.} ReID accuracy reduction, we find an opposite trend for mAP/Rank1 improvements on the two largest datasets MSMT17 and LaST (`\textcolor{red}{red}' in Table~\ref{tab:table1}). This reveals that the larger of unlabeled dataset, the more superior of our method. We analyze such gains come from two aspects: 1). less meta-training data gets more reliable clustered results; 2). the prototype-referenced polishment with intra- and inter-identity constraints promotes the discriminative ReID feature learning.

% \noindent\textbf{Study on Noisy Labeling.} 

% In order to validate the necessity of alleviating noisy labeling defect for U-ReID

% In order to validate the necessity of alleviating noisy labeling defect for U-ReID

\noindent\textbf{Necessity of MCL.} Someone may think of directly splitting a large-scale dataset into multiple small subsets to do clustering-based U-ReID sequentially. This is also the most straightforward solution to handle the computational issue we focused. To study its feasibility, we deliberately design a scheme named \emph{Naive Splitting Training}, where multiple subsets picked from one single large data set are \emph{sequentially} used for \emph{clustering$\rightarrow$labeling$\rightarrow$training}. \emph{Naive Splitting Training} also could save memory cost due to its subset-wise clustering, but this operation also inadvertently enlarges the negative effect of time consuming and noisy labeling. As shown in Table~\ref{tab:table2}, two \emph{Naive Splitting Training} schemes of using \emph{50\%/25\%} subset as training unit, are inferior to \emph{MCL} by 16.0\%/9.7\% in mAP on MSMT17, which reveals two facts that 1) naively splitting the holistic large-scale dataset for sequential training is not optimal, and 2) MCL is necessary and more superior.

% ). We can see MCL consistently outperforms \emph{Naive Splitting Training} on the different datasets and settings

% Table generated by Excel2LaTeX from sheet 'paper_ablation_2'
% \vspace{-8mm}
\begin{table}
  \centering
  \footnotesize
  \caption{Study on the necessity of our MCL. For the scheme of \emph{Naive Splitting Training}, several subsets picked from one single large data set are \emph{sequentially} used for \emph{clustering$\rightarrow$labeling$\rightarrow$training} (same with phase-1 in MCL).}
  \setlength{\tabcolsep}{1.0mm}{
  \tablestyle{4pt}{1.4}
    \begin{tabular}{c|ccccc}
    \toprule
    \multicolumn{2}{c}{\multirow{2}[4]{*}{Methods}} & \multicolumn{2}{c}{Market1501} & \multicolumn{2}{c}{MSMT17} \\
\cmidrule{3-6}    \multicolumn{2}{c}{} & mAP   & Rank1 & mAP   & Rank1 \\
    \midrule
    \multirow{2}[1]{*}{50\%} & Naive Splitting Training & 73.6  & 82.2  & 22.2  & 43.9 \\
          & MCL   & 82.9 (↑{\tiny\tcr{9.3}}) & 92.7 (↑{\tiny\tcr{10.5}}) & 38.2 (↑{\tiny\tcr{16.0}}) & 66.5 (↑{\tiny\tcr{22.6}}) \\
    \hline
    \multirow{2}[1]{*}{25\%} & Naive Splitting Training & 68.4  & 74.1  & 16.2  & 36.5 \\
          & MCL   & 75.4 (↑{\tiny\tcr{7.0}}) & 89.3 (↑{\tiny\tcr{15.2}}) & 25.9 (↑{\tiny\tcr{9.7}}) & 53.4 (↑{\tiny\tcr{16.9}}) \\
    \bottomrule
    \end{tabular}}%
    \vspace{-2mm}
  \label{tab:table2}%
\end{table}%

% alleviating noisy labeling defect is necessary for U-ReID, especially for the subset-wise clustering techniques. 

% Two losses study based on MCL. Besides, adding them on basic baseline, still obtains gains. \emph{MCL} with the proposed two losses outperforms those without anyone constraint. 

\subsection{Study on \emph{Mixed} Large-scale Datasets}

% \noindent\textbf{Meta-Prototype outperforms All-in clustering in more larger dataset}  and prove its correctness

% \noindent\textbf{Larger Size, More Problematic of \emph{All-in}, More Superior of MCL.} 
% mimic the real-world scenario to 

% by mixing multiple public ReID datasets, where a model is trained on \emph{PX+Ma+Duke+MT} or \emph{PX+Ma+Duke+MT+LS} while tested 

% study on the larger \emph{mixed} ReID datasets. Specifically, we

As discussed in Sec.~\ref{sec:table1} and Table~\ref{tab:table1}, the larger size of unlabeled dataset, the more superior of MCL. To fully study this point, we further construct two \emph{mixed} large-scale training datasets \emph{PX+Ma+MT} and \emph{PX+Ma+MT+LS}, and evaluate models on the unseen test set of MSMT17 (\emph{MT}). {Note that, we originally planned to perform such group of experiments on the larger realistic ReID datasets, but which is limited by the truth that most large-scale realistic ReID datasets (\egno, Person30K~\cite{bai2021person30k}, FastHuman~\cite{he2021semi}) have not fully released.} As shown in Table~\ref{tab:large}, we can get two observations: 1). the scheme of \emph{All} on the largest dataset \emph{PX+Ma+MT+LS} is failed to be directly clustered/trained due to the computing pressure. 2). MCL outperforms \emph{All} by 2.3\% in mAP under 50\% on \emph{PX+Ma+MT}$\rightarrow$\emph{MT} 3). {MCL performs better on \emph{PX+Ma+MT} than that on \emph{PX+Ma+MT+LS}, which may be due to the style/domain gap between LaST~\cite{shu2021large} and other ReID datasets.}

\vspace{-1mm}
\begin{table}
  \centering
  \footnotesize
  \caption{Study on \emph{mixed} large-scale datasets, where \emph{PX, Ma, MT, LS} denotes \emph{PersonX, Market1501, MSMT17, LaST}. Note that, in the experimental environment with four 16GB Tesla V100 GPUs, the scheme of \emph{All} on \emph{PX+Ma+MT+LS} is failed to be directly clustered/trained due to the computing pressure.}
  \setlength{\tabcolsep}{2.4mm}{
  \tablestyle{8pt}{1.2}
    \begin{tabular}{c|c|cccc}
    \toprule
    \multirow{2}[1]{*}{Train Datasets} & \multirow{2}[1]{*}{Methods} & \multicolumn{4}{c}{Test: MT} \\
    \cline{3-6}      &       & mAP   & Rank1 & M (MB)     & T (s) \\
    \hline
    MT    & All   & 33.4  & 62.9  & 6251.5 & 118.3 \\
    \hline
    \multirow{3}[0]{*}{PX+Ma+MT} & All   & 29.6  & 56.3  & 29788.7 & 244.8 \\
          & MCL, 50\% & \tcr{31.9}  & \tcr{59.3}  & \tcr{7698.3} & \tcr{82.6} \\
          & MCL, 25\%   & 23.1  & 49.6      & 1399.0 & 35.9  \\
    \hline
    \multirow{3}[2]{*}{PX+Ma+MT+LS} & All   & -- & --      & --      & -- \\
          & MCL, 50\% & \tcr{25.5}  & \tcr{49.9}  & \tcr{21207.8} & \tcr{323.3} \\
          & MCL, 25\% & 17.1 & 39.5      & 5126.9  & 107.5 \\
    \bottomrule
    \end{tabular}}%
    \vspace{-1mm}
  \label{tab:large}%
\end{table}%

\subsection{Ablation Study}\label{sec:ablation}

\noindent\textbf{Influence of Loss Constraints.} We study the effectiveness of the proposed siamese consistency loss $\mathcal{L}_{sc}$ and soft-weighted triplet loss $\mathcal{L}_{tri}^{sw}$ in Table~\ref{tab:loss}. We see that \emph{MCL} outperforms \emph{MCL w/o $\mathcal{L}_{sc}$} by \textbf{4.0\%}/\textbf{3.2\%} in mAP for 50\%/25\% settings on Market1501. When replace the soft-weighted triplet loss $\mathcal{L}_{tri}^{sw}$ with the basic triplet loss version~\cite{hermans2017defense}, the scheme of \emph{MCL w/o $\mathcal{L}_{tri}^{sw}$} is inferior to \emph{MCL} by \textbf{5.6\%}/\textbf{4.4\%} in mAP for 50\%/25\% settings on Market1501. Such two constraints facilitate the pseudo label denoising via promising intra-identity consistency and inter-identity correlation. In addition, they are complementary and both vital to MCL, jointly resulting in a superior performance. 
%  and also discard $\mathcal{L}_{sc}$
\begin{table*}[htbp]\centering
    \caption{Ablation study for meta clustering learning (MCL).}
    \vspace{-2mm}
	% subfloat ############
	\subfloat[Study on the two loss constraints.\label{tab:loss}]{
		\tablestyle{8pt}{1.25}
    \begin{tabular}{c|ccccc}
    \toprule
    \multicolumn{2}{c}{\multirow{2}[4]{*}{Methods}} & \multicolumn{2}{c}{Market1501} & \multicolumn{2}{c}{MSMT17} \\
\cmidrule{3-6}    \multicolumn{2}{c}{} & mAP   & Rank1 & mAP   & Rank1 \\
    \midrule
    \multirow{3}[2]{*}{50\%} & MCL w/o $\mathcal{L}_{sc}$ & 78.9  & 88.8  & 35.1  & 62.8 \\
          & MCL w/o $\mathcal{L}_{tri}^{sw}$ & 77.3  & 86.4  & 33.4  & 60.7 \\
          & MCL   & \tcr{82.9}  & \tcr{92.7}  & \tcr{38.2}  & \tcr{66.5} \\
    \hline
    \multirow{3}[2]{*}{25\%} & MCL w/o $\mathcal{L}_{sc}$ & 72.2  & 84.8  & 20.4  & 44.5 \\
          & MCL w/o $\mathcal{L}_{tri}^{sw}$ & 71.0  & 84.1  & 17.6 & 38.1  \\
          & MCL   & \tcr{75.4}  & \tcr{89.3}  & \tcr{25.9}  & \tcr{53.4} \\
    \bottomrule
    \end{tabular}}\hspace{9mm}
	% subfloat ############
	\subfloat[Study on the data split strategies. \label{tab:data_split}]{
		\tablestyle{8pt}{1.25}
            \begin{tabular}{c|ccccc}
    \toprule
    \multicolumn{2}{c}{\multirow{2}[4]{*}{Methods}} & \multicolumn{2}{c}{Market1501} & \multicolumn{2}{c}{MSMT17} \\
\cmidrule{3-6}    \multicolumn{2}{c}{} & mAP   & Rank1 & mAP   & Rank1 \\
    \midrule
    \multirow{3}[2]{*}{50\%} & MCL\_fixed & 71.5  & 87.8  & 21.2  & 44.0 \\
          & MCL\_same & 80.8  & 92.0  & 36.5  & 63.8 \\
          & MCL   & \tcr{82.9}  & \tcr{92.7}  & \tcr{38.2}  & \tcr{66.5} \\
    \hline
    \multirow{3}[2]{*}{25\%} & MCL\_fixed & 37.8  & 61.0  & 20.0  & 43.7 \\
          & MCL\_same & 59.8  & 80.3  & 14.4  & 34.3  \\
          & MCL   & \tcr{75.4}  & \tcr{89.3}  & \tcr{25.9}  & \tcr{53.4} \\
    \bottomrule
    \end{tabular}}\hspace{3mm}
	\vspace{-3mm}
	\label{tab:ablation}
\end{table*}
%##################################################################################################

% % Table generated by Excel2LaTeX from sheet 'paper_ablation_2'
% \begin{table}[htbp]
%   \centering
%   \scriptsize
%   \caption{Effectiveness of the proposed two loss constraints, siamese consistency loss $\mathcal{L}_{sc}$ and soft-weighted triplet loss $\mathcal{L}_{tri}^{sw}$. Note that, the scheme of MCL w/o $\mathcal{L}_{tri}^{sw}$ means we only use the most commonly-used basic triplet loss~\cite{hermans2017defense}.}
%   \setlength{\tabcolsep}{2.8mm}{
%     \begin{tabular}{c|ccccc}
%     \toprule
%     \multicolumn{2}{c}{\multirow{2}[4]{*}{Methods}} & \multicolumn{2}{c}{Market1501} & \multicolumn{2}{c}{MSMT17} \\
% \cmidrule{3-6}    \multicolumn{2}{c}{} & mAP   & Rank1 & mAP   & Rank1 \\
%     \midrule
%     \multirow{3}[2]{*}{50\%} & MCL w/o $\mathcal{L}_{sc}$ & 78.9  & 88.8  & 35.1  & 62.8 \\
%           & MCL w/o $\mathcal{L}_{tri}^{sw}$ & 77.3  & 86.4  & 33.4  & 60.7 \\
%           & MCL   & \tcr{82.9}  & \tcr{92.7}  & \tcr{38.2}  & \tcr{66.5} \\
%     \hline
%     \multirow{3}[2]{*}{25\%} & MCL w/o $\mathcal{L}_{sc}$ & 72.2  & 84.8  & 20.4  & 44.5 \\
%           & MCL w/o $\mathcal{L}_{tri}^{sw}$ & 71.0  & 84.1  & 17.6 & 38.1  \\
%           & MCL   & \tcr{75.4}  & \tcr{89.3}  & \tcr{25.9}  & \tcr{53.4} \\
%     \bottomrule
%     \end{tabular}}%
%     \vspace{-4mm}
%   \label{tab:loss}%
% \end{table}%

\noindent\textbf{Influence of Data Split.} As we described before Sec.~\ref{sec:phase1}, given an unlabeled dataset $\sX$, we use an random and uniform split strategy to divide the samples into \emph{meta training subset} $\sX_{1}$ and \emph{the rest subsets} $\{\sX_{2},\sX_{3},...,\sX_{N}\}$. Such split is performed before each training epoch. And, the label spaces for different subsets $\{\sX_{1},\sX_{2},\sX_{3},...,\sX_{N}\}$ are same-size but non-overlapping. Here we study on the influence of different split designs. In Table~\ref{tab:data_split}, the scheme of \emph{MCL\_fixed} means we only conduct the data split once at beginning and fix the split results during the training. \emph{MCL\_same} means the scheme where all subsets share the same label space. We can observe that \emph{MCL\_fixed} is inferior to \emph{MCL} by 11.4\%/37.6\% in mAP under 50\%/25\% on Market1501, and \emph{MCL\_same} is inferior to \emph{MCL} by 15.6\%/11.5\% in mAP under 25\% on Market1501/MSMT17. We analyze that: 1). re-splitting dataset before each epoch plays a role of data re-organization, like the mechanism behind cross-validation~\cite{kohavi1995study}, which avoids over-fitting and extremely cases, increasing robustness of MCL. 2). non-overlapped label spaces increase the diversity of training data, like a data augmentation, promoting the discriminative ReID representations learning. Such design brings obvious improvements especially when using less meta-training data. For example, \emph{MCL} outperforms \emph{MCL\_same} by only 2.1\%/1.7\% in mAP under 50\%, but by 15.6\%/11.5\% under 25\%. {More analytic and ablated results (including limitation discussions) are presented in \textbf{Supplementary}.}

\subsection{Visual Results and Insights}

%  This group of experiments are conducted on MSMT17 and LaST datasets, our scheme is \emph{MCL, 50\%}. 

\begin{figure}
  \centerline{\includegraphics[width=.98\linewidth]{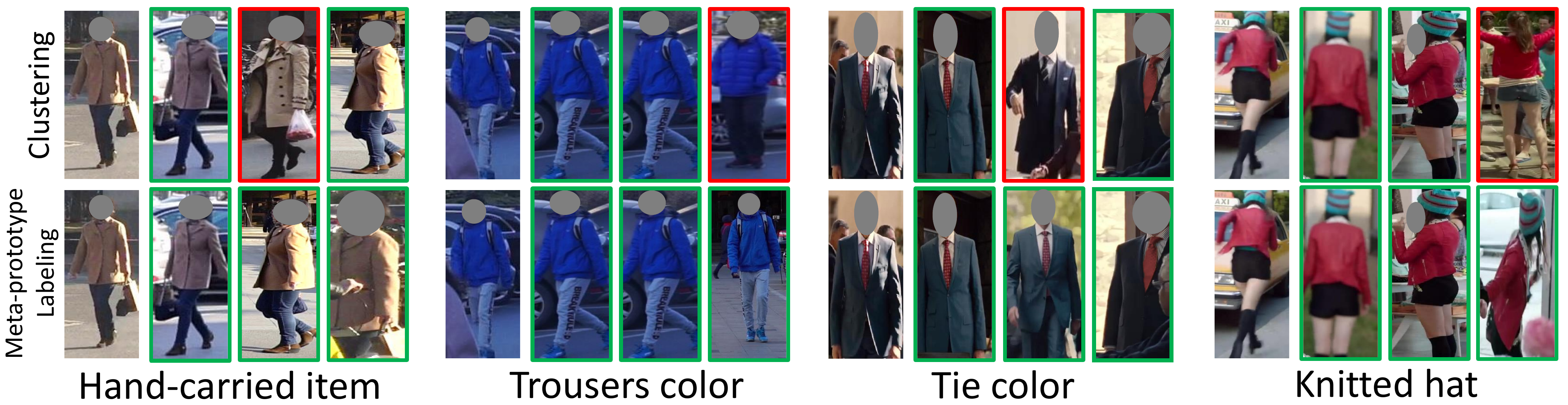}}
  \vspace{-2mm}
    \caption{Visual results of the same pseudo-labeled images on large-scale MSMT17 (two groups on the left) and LaST (two groups on the right), mined by using the general clustering technique and our meta-prototype referenced labeling. Green boxes denote correct results while red boxes denote false results. Important fine-grained feature clues are highlighted below each image pair. All faces in the images are masked for anonymization.}
\label{fig:visualization}
\vspace{-3.5mm}
\end{figure}

\noindent\textbf{Visualization on Pseudo Labeling.} To further show the proposed prototype-referenced labeling in MCL is superior to the general clustering, we compare these two pseudo-labeling methods by showing the same pseudo-labeled images (\ieno, the positive pairs) in Figure~\ref{fig:visualization}. Our scheme is \emph{MCL, 50\%}. We observe that: 1). for the general clustering, the grouped entries share the global visual similar appearance. This is not reliable enough. For example, in the most left pair of Figure~\ref{fig:visualization}, the two women are dressed very similarly, the only local discriminative clue is they take different items in their hands. 2). the proposed meta-prototype referenced labeling has the capability of discovering fine-grained discriminative clues (bottom in Figure~\ref{fig:visualization}) due to the usage of relative comparative characteristic among samples. This also explains why MCL outperforms the baseline scheme of \emph{All} even with less data for clustering to some extent.

% \vspace{-5.5mm}
\begin{figure}
  \centerline{\includegraphics[width=0.99\linewidth]{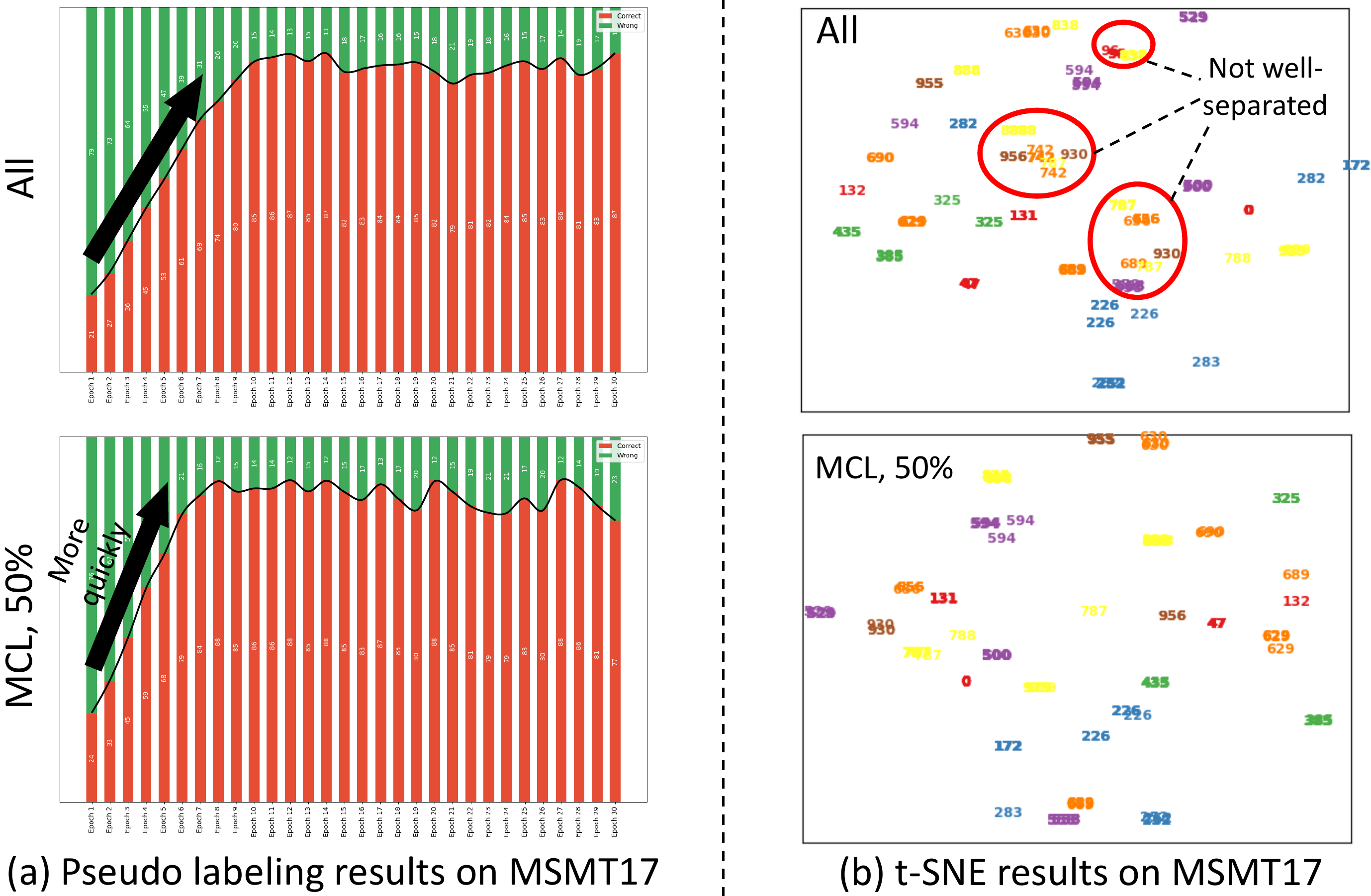}}
  \vspace{-2mm}
    \caption{(a) Visualization of the percentages of correctly (red) and wrongly (green) pseudo labeling on MSMT17 as training goes on. (b) Visualization of t-SNE distributions on MSMT17. Different colors and digits represent different identities.}
\label{fig:vis_stacked_hist_tsne}
\vspace{-4mm}
\end{figure}

\begin{table*}[htbp]\centering
    
    \caption{Comparison with state-of-the-art methods on the unsupervised ReID, including purely unsupervised methods and unsupervised domain adaptation (UDA) methods. ``None'' represents the former, and other value represents the source-domain dataset in UDA method.}
    \vspace{-2mm}
	% subfloat ############
	\subfloat[Experiments on Market1501.\label{tab:market1501}]{
		\tablestyle{8pt}{1.1}
    \begin{tabular}{l|c|cccc}
        \hline
        \multicolumn{1}{c|}{\multirow{2}*{Methods}} &
        \multicolumn{5}{c}{Market1501}\\
        \cline{2-6}
        &source & mAP & Rank1 & Rank5 & Rank10 \\ 
        \hline
        BUC~\cite{lin2019bottom}  & None & 38.3 & 66.2 & 79.6 & 84.5 \\
        UGA~\cite{wu2019unsupervised}  & None & 70.3 & 87.2 & - & - \\        
        SSL~\cite{lin2020unsupervised}  & None & 37.8 & 71.7 & 83.8 & 87.4 \\
        MMCL~\cite{wang2020unsupervised}  & None & 45.5 & 80.3 & 89.4 & 92.3 \\
        % MMCL~\cite{wang2020unsupervised} (CVPR'20) & Duke & 60.4 & 84.4 & 92.8 & 95.0 \\
        HCT~\cite{zeng2020hierarchical}  & None & 56.4 & 80.0 & 91.6 & 95.2 \\
        DG-Net~\cite{zou2020joint}  & MT & 64.6 & 83.1 & 91.5 & 94.3 \\
        CycAs~\cite{wang2020cycas}  & None & 64.8 & 84.8 & - & - \\
        % AD-Cluster++~\cite{zhai2020ad} (CVPR'20) & Duke & 68.3 & 86.7 & 94.4 & 96.5 \\
        
        MMT~\cite{ge2020mutual} & MT & 75.6 & 89.3 & 95.8 & 97.5 \\
        SPCL~\cite{ge2020self}  & None & 73.1 & 88.1 & 95.1 & 97.0 \\
        SPCL~\cite{ge2020self}  & MT & 77.5 & 89.7 & 96.1 & 97.6 \\

        MPRD~\cite{ji2021meta}  & None & 51.1 & 83.0 & 91.3 & 93.6  \\
        
        ICE~\cite{chen2021ice}  & None & 79.5 & 92.0 & 97.0 & 98.1 \\ 
        
        HCD~\cite{zheng2021online}  & MT & 80.2 & 91.4 & - & - \\ 
        
        Cluster~\cite{dai2021cluster} & None & 82.6 & {\bf 93.0} & 97.0 & 98.1 \\
        \hline
        {\bf MCL, 50\%} & None & {\bf 82.9} & {92.7} & {\bf 97.6} & {\bf 98.7} \\
        
        % {\bf Ours/IBN-ResNet-50} & None & {\bf 84.1} & {\bf 93.2} & {\bf 97.6} & {\bf 98.2} \\
        \hline
        \end{tabular}}\hspace{10mm}
	% subfloat ############
	\subfloat[Experiments on MSMT17. \label{tab:msmt17}]{
		\tablestyle{8pt}{1.24}
            \begin{tabular}{l|c|cccc}
                \hline
                \multicolumn{1}{c|}{\multirow{2}*{Methods}} &
                \multicolumn{5}{c}{MSMT17}\\
                \cline{2-6}
                &source & mAP & Rank1 & Rank5 & Rank10 \\ 
            \hline
            TAUDL~\cite{li2018unsupervised}  & None & 12.5 & 28.4 & - & - \\
            % ECN~\cite{zhong2019invariance} (CVPR'19) & Duke & 10.2 & 30.2 & 41.5 & 46.8\\
            MMCL~\cite{wang2020unsupervised}  & None & 11.2 & 35.4 & 44.8 & 49.8 \\
            
            UTAL~\cite{li2019unsupervised}  & None & 13.1 & 31.4 & - & - \\
            
            UGA~\cite{wu2019unsupervised}  & None & 21.7 & 49.5 & - & - \\
            MMT~\cite{ge2020mutual}  & Ma & 24.0 & 50.1 & 63.5 & 69.3 \\
            CycAs~\cite{wang2020cycas}  & None & 26.7 & 50.1 & - & - \\
            SPCL~\cite{ge2020self}  & None & 19.1 & 42.3 & 55.6 & 61.2 \\
            SPCL~\cite{ge2020self}  & Ma & 26.8 & 53.7 & 65.0 & 69.8 \\
        
            MPRD~\cite{ji2021meta}  & None & 14.6 & 37.7 & 51.3 & 57.1 \\ 
            
            HCD~\cite{zheng2021online}  & None & 26.9 & 53.7 & 65.3 & 70.2 \\
            
            ICE~\cite{chen2021ice}  & None & 29.8 & 59.0 & 71.7 & 77.0 \\
            
            Cluster~\cite{dai2021cluster}  & None & 33.3 & 63.3 & 73.7 & 77.8 \\
            \hline
            {\bf MCL, 50\%} & None & {\bf 38.2} & {\bf 66.5} & {\bf 75.2} & {\bf 79.7} \\ 
            \hline
            \end{tabular}}\hspace{3mm}
	\vspace{-1mm}
	\label{tab:sota}
\end{table*}
%##################################################################################################

Moreover, we also count the proportions
of correctly and wrongly clustering persons into the same category on MSMT17 in Figure~\ref{fig:vis_stacked_hist_tsne}~(a), we can see that \emph{MCL, 50\%} could achieve a better identity grouping performance more quickly compared to the baseline scheme of \emph{All}.

% different person categories being clustered into 

% show the cluster feature representation As shown in Figure 3
% (b), the instance-level memory averages all instance feature
% vectors to represent the cluster feature. However, in USL re-
% ID, the pseudo label generation stage would inevitably introduce
% the outlier instances. In Figure 6, we count the proportions
% of different real categories being clustered into the
% same category on the Market-1501 dataset. It shows there
% still around 20% noisy instances when model training in finished.

% \begin{figure}
%   \centerline{\includegraphics[width=1\linewidth]{figures/vis_tsne_v2.pdf}}
%   \vspace{-2mm}
%     \caption{Visualization of t-SNE distributions on (a) Market1501 and (b) MSMT17. Compared with \emph{All} scheme, the features of different identities (marked with different colors) are better clearly separated for our scheme of \emph{MCL, 50\%}. Identities that are not well separated are circled in red.}
% \label{fig:vis_tsne}
% \vspace{-5mm}
% \end{figure}

\noindent\textbf{Visualization of Feature Distributions.} In Figure \ref{fig:vis_stacked_hist_tsne}~(b), we visualize the distributions of the features using t-SNE~\cite{van2008visualizing} on MSMT17. We compare the feature distribution with the baseline scheme of \emph{All}, and observe that the features of different identities are better clearly separated for our scheme \emph{MCL, 50\%}, which demonstrates our learned ReID representations are more discriminative.

% , the number means the ID

\subsection{Comparison with State-of-the-arts}

Although this work is the first attempt to achieve the unsupervised ReID learning while considering the computational cost savings, we also compare MCL to the state-of-the-art U-ReID methods that without considering resource limitations. From Table~\ref{tab:sota}, we can see that \emph{MCL, 50\%} using only 50\% unlabeled data for meta-clustering achieves a comparable U-ReID performance compared to SOTA methods, and even outperforms the second best ClusterContrast~\cite{dai2021cluster} by 4.9\% in mAP on the large-scale MSMT17. In short, MCL is capable of achieving a good trade-off between U-ReID performance and computational costs.

% \subsection{Hyper-parameter Analysis}

% \noindent\textbf{Influence of the Loss Balance.} 
% As we described in the manuscript, we design a siamese consistency loss $\mathcal{L}_{sc}$ and a soft-weighted triplet loss $\mathcal{L}_{tri}^{sw}$ in the second polishment phase of meta-clustering learning (MCL) to alleviate the noisy pseudo label issue, which process can be formulated by $\mathcal{L}_{phase2} = \mathcal{L}_{sc}+\lambda*\mathcal{L}_{tri}^{sw}$, the hyper-parameter $\lambda$ is used to balance the importance between the siamese consistency loss $\mathcal{L}_{sc}$ and the soft-weighted triplet loss $\mathcal{L}_{tri}^{sw}$. For $\lambda$, we initially set it to 1.0, and then coarsely determine it based on the corresponding loss values and its gradients observed during the training. The decision principle is to set its value to make the loss value/gradient lie in a similar range. Grid search within a small range of the derived $\lambda$ is further employed to get better parameter. Actually, we observed the final performance is not very sensitive to this hyper-parameter, we experimentally set $\lambda=1.0$ in the end. 

% The similar hyper-parameter strategy is also used to experimentally decide the agreement threshold $A$ and momentum coefficient $m$.

\noindent\textbf{Influence of Clustering Hyper-parameters.} 
As discussed in implementation, we use DBScan and Jaccard distance~\cite{zhong2017re} for first-phase training to cluster with $k$ nearest neighbors ($k$=30) following~\cite{ge2020self,dai2021cluster}. For DBScan, the maximum distance $d$ between two samples is set as 0.4 for market1501, 0.7 for other datasets, and the minimal number of neighbors in a core point (denoted as $n$) is all set as 4. Here we analyze the influence of these parameters in Figure~\ref{fig:para_dbscan}, and conclude that the proposed large-scale unsupervised ReID training method of MCL is robust and stable enough to achieve relatively satisfactory performance with variant hyper-parameters.

% \begin{figure}
%   \centerline{\includegraphics[width=1.0\linewidth]{figures/param_dbscan.pdf}}
%   \vspace{-1mm}
%     \caption{Hyper-parameters exploration for the clustering algorithm of DBScan, where we conduct experiments on Market1501 and MSMT17 with the scheme of \emph{MCL, 50\%}.}
% \label{fig:para_dbscan}
% \end{figure}

\section{Conclusion}

In this paper, we take the first attempt to explore a resource-friendly purely unsupervised person ReID framework, which effectively learns discriminative representations while considering the computational costs. A new concept of meta clustering learning  (MCL) is introduced to perform clustering-based ReID training on \emph{partial} unlabeled data, saving the required computing resources. For \emph{the rest data}, we leverage the learned prototypes obtained before as proxy annotator to pseudo-label them. Based on the generated soft pseudo labels, we then polish model with two well-designed losses that take intra- and inter-identity constraints into account for alleviating noisy labels. MCL achieves a SOTA performance on unsupervised ReID, and could also flexibly meet the computing budgets in practice.

\section{Acknowledgements}

This work was supported in part by NSFC under Grant U1908209, 62021001, the National Key Research and Development Program of China 2018AAA0101400, and NUS Faculty Research Committee Grant (WBS:A-0009440-00-00).

\appendix
  \renewcommand\thesection{\arabic{section}}

\vspace{4mm}
%\Huge
%\huge
%\LARGE
%\Large
\noindent{\LARGE \textbf{Supplementary}}
% \vspace{2mm}

%%%%%%%%% BODY TEXT

\section{Limitations of Meta Cluster Learning}

% Table generated by Excel2LaTeX from sheet 'Sheet3'
\begin{table*}[htbp]
  \centering
  \footnotesize
  \caption{Memory\&Time Cost \textbf{\emph{vs.}} Unsupervised ReID Performance (\%). In which, $M (MB)$, $T (s)$ denotes the \emph{memory cost}, \emph{time cost} of performing clustering \textbf{once} in training, where `s' means `second'. $T (h)$ denotes the total training time where `h' means `hour'. We compare several MCL variants to baseline (\emph{All}) by using \emph{50\%}, \emph{33\%}, \emph{25\%}, and \emph{20\%} data randomly selected from the entire unlabeled dataset as meta-training subset $\sX_{1}$. For the smallest dataset PersonX~\cite{sun2019dissecting}, it is not necessary to do experiments with too harsh computational requirements (\egno, \emph{33\%}, \emph{25\%}, \emph{20\%}). We can see that the larger size of unlabeled dataset, the more superior of our method (red). Note that, the DukeMTMC-ReID dataset~\cite{ristani2016performance} has been taken down and thus not used in our experiment, we just use PersonX~\cite{sun2019dissecting}, Market1501~\cite{zheng2015scalable}, MSMT17~\cite{wei2018person}, and LaST~\cite{shu2021large} for experiments.}
  \renewcommand\arraystretch{1.5}
  \vspace{-2mm}
  \setlength{\tabcolsep}{1.3mm}{
    \begin{tabular}{cccccc|ccccc|ccccc|ccccc}
    \toprule
    \multirow{2}[2]{*}{Methods} & \multicolumn{5}{c|}{PersonX (9.8k imgs, 410 IDs)} & \multicolumn{5}{c|}{Market1501 (12.9k imgs, 751 IDs)} & \multicolumn{5}{c|}{MSMT17 (32.6k imgs, 1041 IDs)}   & \multicolumn{5}{c}{LaST (71.2k imgs, 5000 IDs)} \\
\cmidrule{2-21}          & mAP   & Rank1 & M (MB)    & T (s)  & T (h)     & mAP   & Rank1 & M (MB)    & T (s)    & T (h)  & mAP   & Rank1 & M (MB)    & T (s)  & T (h)    & mAP   & Rank1 & M (MB)    & T (s) & T (h)  \\
    \hline
    
    All   & 88.5  & 95.8  & 822.3 & 30.0 & 2.7 & 83.3  & 93.0  & 876.3 & 34.3 & 2.9 & 33.4  & 62.9  & 6251.5 & 118.3 & 9.3 & 19.8  & 74.0  & 22398.5 & 494.8 & 42.0 \\
    
    50\%  & \textcolor[rgb]{0,  0, 1}{79.0}  & \textcolor[rgb]{0,  0, 1}{93.5}  & \textcolor[rgb]{0,  0, 1}{412.6} & \textcolor[rgb]{0,  0, 1}{13.1} & \textcolor[rgb]{0,  0, 1}{2.2} & 82.9  & 92.7  & 348.6 & 10.8 & 2.4 & \textcolor[rgb]{ 1,  0,  0}{38.2} & \textcolor[rgb]{ 1,  0,  0}{66.5} & \textcolor[rgb]{ 1,  0,  0}{1761.3} & \textcolor[rgb]{ 1,  0,  0}{31.1} & \textcolor[rgb]{ 1,  0,  0}{4.6} & \textcolor[rgb]{ 1,  0,  0}{20.0} & \textcolor[rgb]{ 1,  0,  0}{74.9} & \textcolor[rgb]{ 1,  0,  0}{5779.6} & \textcolor[rgb]{ 1,  0,  0}{121.2} & \textcolor[rgb]{ 1,  0,  0}{20.0}\\
    
    33\%  & --      & --      & --      & -- & -- & 79.6  & 91.9  & 287.5 & 7.0  & 2.2  & 31.5  & 57.4  & 889.1 & 18.2 & 3.8 & \textcolor[rgb]{ 1,  0,  0}{22.7}  & \textcolor[rgb]{ 1,  0,  0}{75.0}  & \textcolor[rgb]{ 1,  0,  0}{2688.2} & \textcolor[rgb]{ 1,  0,  0}{70.8} & \textcolor[rgb]{ 1,  0,  0}{14.0}\\
    
    25\%  & --      & --      & --      & -- & -- & 75.4  & 89.3  & 235.1 & 5.4  & 2.0  & 25.9  & 53.4  & 556.9 & 13.4 & 3.0 & 17.2  & 69.0  & 1564.2 & 44.7 & 9.0 \\
    
    20\%  & --      & --      & --      & -- & -- &  41.3  & 61.3  & 141.4 & 4.6  & 1.9   & 20.1  & 47.4  & 394.6 & 10.3 & 2.6 & 15.8  & 56.0  & 1066.0 & 38.4 & 7.0\\
    
    \bottomrule
    \end{tabular}}%
    \vspace{-1mm}
  \label{tab:table1}%
\end{table*}%

\noindent\textbf{MCL Cannot Work Well with Too Small Meta-training Data $\sX_{1}$.} As we described in the manuscript, MCL costs less computing resources in clustering by only using a meta-training subset $\sX_{1}$ in the first phase for training. Intuitively, if such meta-training subset $\sX_{1}$ is too small, MCL will be seriously affected by the noisy pseudo label issue so that causes an unsatisfactory U-ReID performance. Taking an extreme case as example for illustration, given a person dataset with 10 IDs and 100 person images, if we split it into 10 subsets and just pick up one as $\sX_{1}$, the worst case is that $\sX_{1}$ may only cover a single person, which will directly make the reference-based pseudo labeling and U-ReID failed. The similar phenomenon also can be observed from Table~\ref{tab:table1}, MCL can not work so well with too small meta-training data $\sX_{1}$ especially on small-scale datasets, like \emph{20\%, on PersonX} (marked as blue).

\noindent\textbf{The Relationship between Dataset Size and Optimal Split Number Is Not Clear.} As we claimed in the manuscript, the larger size of unlabeled dataset, the more superior of the proposed MCL. However, it is difficult to give a deterministic relationship between dataset size and split number. As shown in Table~\ref{tab:table1}, the scheme of \emph{All} that directly conduct clustering over full data achieves the best ReID performance on PersonX and Market1501, the scheme of \emph{MCL, 50\%} achieves the best ReID performance on MSMT17, and the scheme of \emph{MCL, 33\%} achieves the best ReID performance on the largest LaST. We conclude that the larger size of unlabeled dataset, MCL might could use more less meta-training data $\sX_{1}$ (\ieno, a big split number) for training to get a satisfactory performance. But, such the relationship that related to the dataset size \emph{vs.} the optimal number of subsets is not clear now. The exploration about this relationship is also limited by the existing/released public person ReID datasets, or said, not so many large-scale person ReID datasets, so this will be left as our future work.

%%%%%%%%%%%%%%%%%%%%%%%%%%%%%%%%%%%%%%%%%%%%%%%%%%%%%%%%%
\section{Meet the Computing Budgets In Practice}

\begin{figure}
  \centerline{\includegraphics[width=1.0\linewidth]{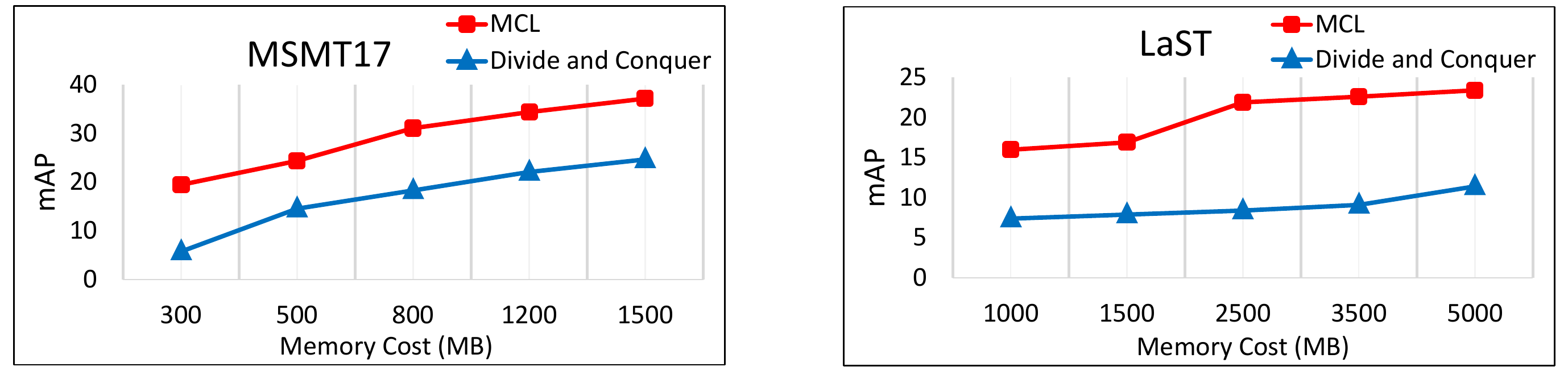}}
  \vspace{-2mm}
    \caption{Given a limited memory budget, MCL could achieve a better discriminative representations learning, outperforming \emph{Naive Splitting Training} scheme on the variable budgets.}
\label{fig:curves}
\vspace{-3mm}
\end{figure}

% via Varying Ratios
% our by-product, minor contribution, 
% Basically, use of coarse-to-fine prediction cascades~\cite{huang2017multi,teerapittayanon2016branchynet,wang2017idk} and dynamic inference approaches~\cite{lin2017runtime,wang2018skipnet} to reduce computational complexity is common in object classification/detection
% tasks. However, their use in person ReID is rare. Overall, there is a need for ReID models can operate efficiently and accurately at continuously varying computational cost constraints. 
Uniquely, our MCL could enable U-ReID to operate at varying computation budgets. As we pointed before, as a by-product of MCL, our trained model could fully leverage the entire unlabeled data set with only a \emph{partial} subset doing clustering. Such \emph{ratio} can be flexibly determined according to the practical computing power. Given a computing budget, we compare the U-ReID performance of MCL with the \emph{Naive Splitting Training} scheme (\ieno, \emph{sequentially} using subsets to meet resource requirements). As the two curves shown in Figure~\ref{fig:curves}, MCL consistently outperforms \emph{Naive Splitting Training} on the two largest datasets MSMT17 and LaST. That is, given a limited budget, MCL could achieve a better discriminative ReID representation learning.

%%%%%%%%%%%%%%%%%%%%%%%%%%%%%%%%%%%%%%%%%%%%%%%%%%%%%%%%%

\section{Hyper-parameter Analysis}

\noindent\textbf{Influence of the Loss Balance.} 
As we described in the manuscript, we design a siamese consistency loss $\mathcal{L}_{sc}$ and a soft-weighted triplet loss $\mathcal{L}_{tri}^{sw}$ in the second polishment phase of meta-clustering learning (MCL) to alleviate the noisy pseudo label issue, which process can be formulated by $\mathcal{L}_{phase2} = \mathcal{L}_{sc}+\lambda*\mathcal{L}_{tri}^{sw}$, the hyper-parameter $\lambda$ is used to balance the importance between the siamese consistency loss $\mathcal{L}_{sc}$ and the soft-weighted triplet loss $\mathcal{L}_{tri}^{sw}$. For $\lambda$, we initially set it to 1.0, and then coarsely determine it based on the corresponding loss values and its gradients observed during the training. The decision principle is to set its value to make the loss value/gradient lie in a similar range. Grid search within a small range of the derived $\lambda$ is further employed to get better parameter. Actually, we observed the final performance is not very sensitive to this hyper-parameter, we experimentally set $\lambda=1.0$ in the end. 

\begin{figure}
  \centerline{\includegraphics[width=1.0\linewidth]{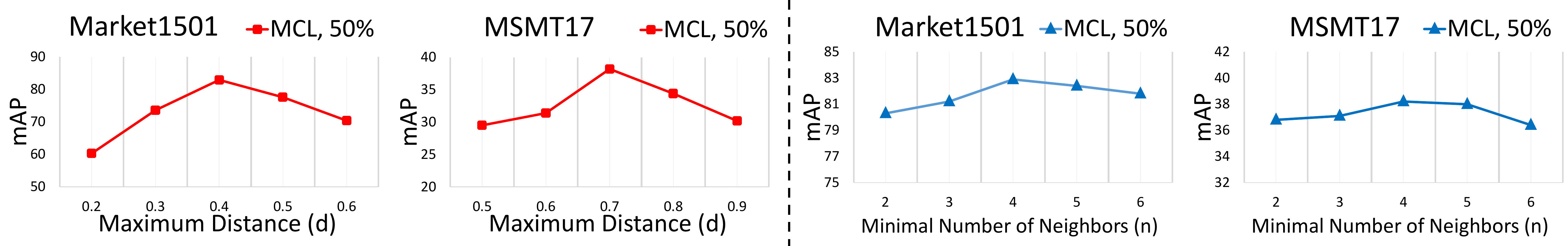}}
  \vspace{-1mm}
    \caption{Hyper-parameters exploration for the clustering algorithm of DBScan, where we conduct experiments on Market1501 and MSMT17 with the scheme of \emph{MCL, 50\%}.}
\label{fig:para_dbscan}
\end{figure}

% 是的，最好是能给出一个dataset size和最优subset数量的规律

% 那是最好的了···wc，但无奈受限于现有数据集啊，我们也不好给，，，好合理，把锅甩给数据集，hahaha

% 对的，dataset size和subset的规律或者是number of pseudo labels和subset的规律

% 1. zizheng: 是的，我看到了。我觉得这个每个epoch的效果很大程度上取决于X1的数据分布。如果X1中sample的样本都很散乱，每个id的都有，那么聚类产生的K值会很大，那么这个epoch过程产生的伪标签数量会很大，感觉标签分布就有点散了；如果X1的数据分布太集中可能也不好。所以师兄每个epoch都会重新采样重新选X1，感觉挺合理的。

% 但我个人是有一种感觉，不知道对不对，就是：把整个数据集分的越开（subset越多），相对的话训练过程中产生的伪标签数会越多

% 所以这种方法在id相对较少的数据集上效果不好，而id多的大数据上效果更好

% 2. 说不定真是这样（瞎猜的）。因为看你的Table 1，随着数据集规模和id数增大，subset相对越多，效果越好。比如id和规模都很小的market1501和personx，subset=1效果最好；稍微大一些的MSMT17，subset=2效果最好；再大的LaST，subset=3效果最好。要是还有更大的数据集，说不定subset=4效果最好。
% 但是感觉凭师兄展示上去的结果还不能支持得到这种结论。

% 是的，最好是能给出一个dataset size和最优subset数量的规律

% 那是最好的了···wc，但无奈受限于现有数据集啊，我们也不好给，，，好合理，把锅甩给数据集，hahaha

% 对的，dataset size和subset的规律或者是number of pseudo labels和subset的规律

% Loss Hyperparameter evaluations.

\section{Social Impact}

\noindent\textbf{Positive.} In this paper, we introduce the Meta Clustering Learning (termed as MCL), a new concept for large-scale person re-identification. To our best knowledge, this paper is the first attempt to develop an efficient unsupervised person ReID training framework to fully leverage the numerous pedestrian surveillance data while taking the computational cost into account. This is very important for both of academic community and industry, and is also valuable and meaningful to bridge the gap between the fast-developing ReID algorithms and practical applications.

This paper also has the potential to provide new insights to the person ReID field and accelerate the development of ReID algorithms. We improve ReID model's learning ability, enabling the models to be trained with limited computing resources in practical applications. Moreover, the prototype-referenced peseudo labeling idea, and the well-designed intra- and inter-identity constraints are conceptually suitable for a wide range of tasks: from person detection, matching, retrieval to fine-grained person tracking, \etcno.

\noindent\textbf{Negative.} Due to the urgent demand of public safety and
increasing number of surveillance cameras, person ReID is imperative in intelligent surveillance systems with significant research impact and practical importance, but this task also might raise questions about the risk of leaking private information. On the other hand, the data collected from the surveillance equipments or downloaded from the internet may violate the privacy of human beings. Therefore, we appeal and encourage further person ReID work to understand and avoid as much as possible the risks of using these pedestrian data. We also encourage research that understands and mitigates the risks arising from surveillance applications. A short-term solution may be developing detection systems. Besides, we recommend researchers to stop the spread of private datasets.

% %%
% %% The acknowledgments section is defined using the "acks" environment
% %% (and NOT an unnumbered section). This ensures the proper
% %% identification of the section in the article metadata, and the
% %% consistent spelling of the heading.
% \begin{acks}
% \end{acks}

\bibliographystyle{ACM-Reference-Format}
\bibliography{sample-base}

% %%
% %% If your work has an appendix, this is the place to put it.
% \appendix









\end{document}